\documentclass{SCIS2021}

\usepackage[english]{babel}
\usepackage{amsthm}

\usepackage[table]{xcolor} 
\definecolor{darkgray}{rgb}{0.7, 0.7, 0.7} 
\definecolor{gray}{rgb}{0.8, 0.8, 0.8}     
\definecolor{lightgray}{rgb}{0.9, 0.9, 0.9} 
\usepackage{multirow}
\usepackage[colorlinks,linkcolor=black]{hyperref}
\newtheoremstyle{exampstyle}
{0.0em} 
{0.0em} 
{} 
{1em} 
{\bfseries} 
{.} 
{1em} 
{} 
\usepackage{balance}
\theoremstyle{exampstyle}

\usepackage{CJKutf8}


\begin{document}
	\ArticleType{REVIEW}
	\SpecialTopic{Artificial Intelligence Innovation in Remote Sensing}
	\Year{2025}
	\Month{}
	\Vol{}
	\No{}
	\DOI{}
	\ArtNo{}
	\ReceiveDate{}
	\ReviseDate{}
	\AcceptDate{}
	\OnlineDate{}

	\title{State Space Models Meet Remote Sensing: A Survey}{State Space Models Meet Remote Sensing: A Survey}

	\author[1]{Qinzhe YANG}{}
        \author[1,2,3]{Chenyang LIU}{}
        \author[4]{Jia XU}{}
        \author[2,3]{Zhenwei SHI}{}
        \author[2,3]{Zhengxia ZOU}{{zhengxiazou@buaa.edu.cn}}

	\AuthorMark{Qinzhe YANG}

		\AuthorCitation{Yang Q, Liu C, Xu J, et al}

	\address[1]{Shen Yuan Honors College, Beihang University, Beijing {\rm 100191}, China}
        \address[2]{Department of Aerospace Intelligent Science and Technology, School of Astronautics, Beihang University, Beijing {\rm 100191}, China}
        \address[3]{State Key Laboratory of Virtual Reality Technology and Systems, Beihang University, Beijing {\rm 100191}, China}
        \address[4]{Qian Xuesen Laboratory of Space Technology, China Academy of Space Technology, Beijing {\rm 100094}, China}

	\abstract{State Space Models (SSMs), designed for long-range modeling, offer linear computational complexity and strong capabilities in capturing long-range dependencies. In the field of remote sensing, SSMs have gained popularity due to their effectiveness in addressing unique challenges such as dense visual predictions, multi-modal remote sensing data, and temporal remote sensing data, which have also yielded significant advancements in customized architectures. This paper presents a comprehensive review of SSM-based approaches in remote sensing, covering most of the relevant studies since SSMs were first introduced to the field. We offer a multi-dimensional analysis examining SSM applications in remote sensing tasks and discussing advancements in architecture design. This paper not only synthesizes the rapid progress in SSM-based research but also identifies key challenges and future opportunities. By providing a detailed perspective, this paper aims to serve as a foundational resource for remote sensing researchers, offering actionable insights to foster further advancements in this evolving domain. We will keep tracing related works at \href{https://github.com/QinzheYang/Awesome-RS-State-Space-Model}{\textcolor{blue}{https://github.com/QinzheYang/Awesome-RS-State-Space-Model}}.}

	
	\maketitle
	\section{Introduction}

    State Space Models (SSMs) are a novel framework introduced to the field of artificial intelligence last year. They achieve linear complexity, making them particularly suitable for tasks involving lengthy sequences \cite{1,2,3,4,5,6}. They have demonstrated advantages in long-sequence modeling tasks over existing models\cite{7,8,9}. Since images can also be unfolded into sequential data in a certain order, this efficiency has led to significant success in the vision domain. SSMs have shown remarkable performance in tasks such as image classification, object detection, and semantic segmentation \cite{10,11,12,13,14,15,16,17,18}.

    In the field of remote sensing, the beginning of 2024 has seen a notable surge of interest in SSMs, fueled by the unique challenges and demands inherent to this domain \cite{19,20,21,22,23,24}. Remote sensing tasks are characterized by dense prediction requirements, multi-modal data types, and the need to process temporal remote sensing data\cite{1,25,26,27,28,29,30,31,32,33,34,35,36,37}. These challenges have necessitated the development of specialized architectures capable of efficiently modeling long sequences while addressing the specific demands of remote sensing applications. 

    SSM, with its efficient long-sequence modeling capabilities, has been widely studied in various tasks to address the unique challenges mentioned above\cite{38,39,40}. For instance, in dense prediction tasks such as semantic segmentation, SSM-based backbones can be computationally attractive because their sequence modeling scales approximately linearly with token length, which may ease modeling large spatial extents. \cite{41,42}. Moreover, remote sensing often involves abundant multi-modal data, such as hyperspectral (HSI), multispectral (MSI), and panchromatic (Pan) imagery. By integrating the spatial and spectral scanning mechanisms, SSMs can establish correlations among these diverse modalities \cite{43,44}. Additionally, when processing multi-temporal remote sensing information, SSMs can sensitively perceive feature differences across different time periods, leading to numerous advancements in the field\cite{45,46,47}. Since 2024, the uniqueness of remote sensing has catalyzed a series of model architectural innovations, including novel scanning strategies, hybrid architectures, and lightweight designs, all aimed at improving the efficiency and effectiveness of SSMs in remote sensing applications. Therefore, a comprehensive review of SSMs in remote sensing is urgently needed.

\begin{figure*}
	\centering
	\includegraphics[width=1\linewidth]{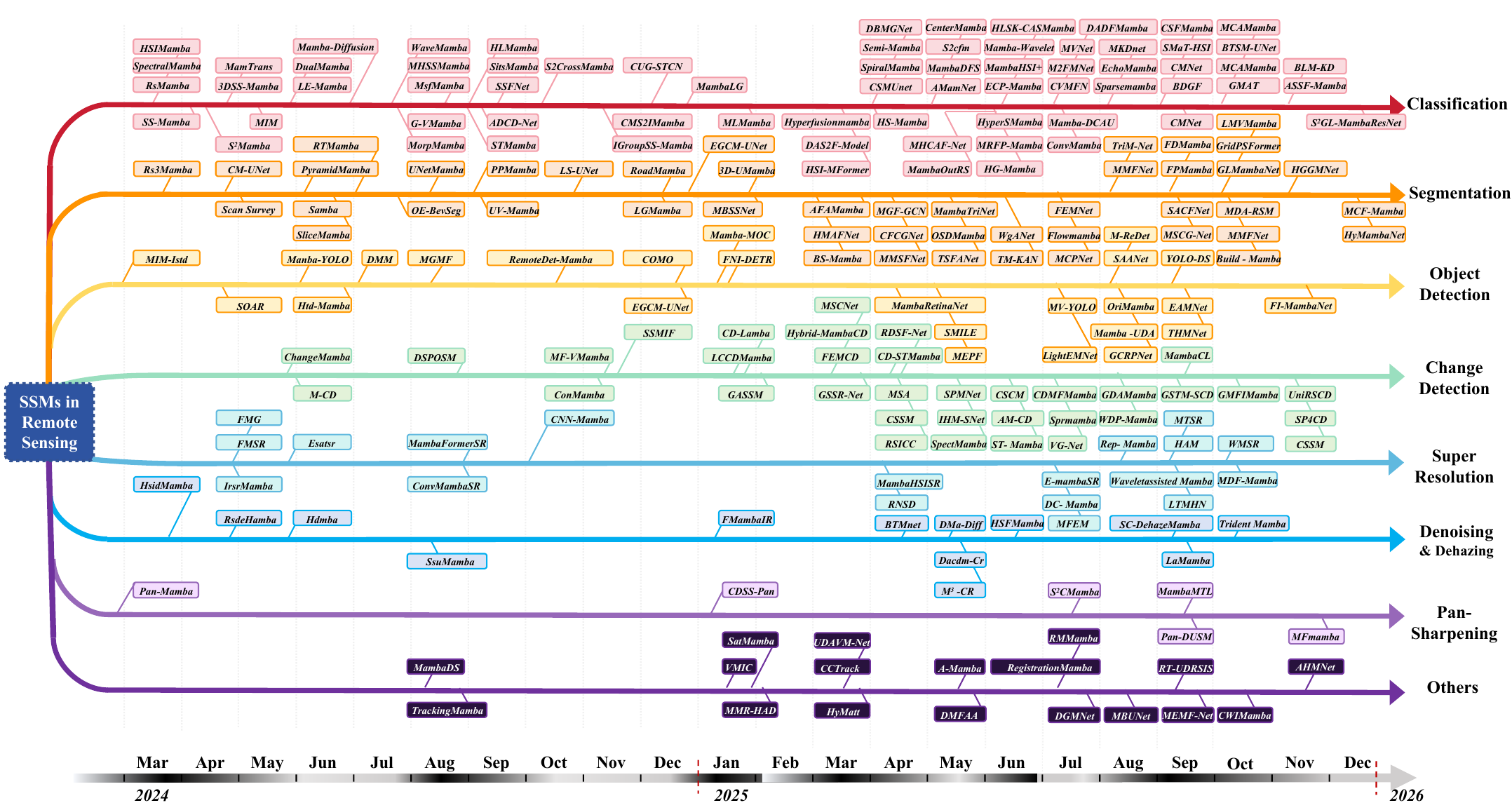}
        \vspace{-4ex}
	\caption{{The recent development of the SSMs in remote sensing since March 2024. It has showcased new research of SSMs on different remote sensing tasks. }}
	\label{fig:development}
\end{figure*}

    Although there are currently some reviews on SSMs, most of them focus on fields such as computer vision \cite{28,29,31,32,33}. However, remote sensing images differ from general images and inherently involve characteristics such as multimodal heterogeneity, frequency domain noise, and long-range dependencies. This necessitates that the design of SSMs takes into account issues such as multimodal fusion, noise removal, and global-local dependency relationships. Many researchers have conducted excellent studies on RS-SSM to address these issues. Therefore, it is essential to conduct a review specifically on SSMs for remote sensing. Recently, Bao \textit{‌et al.} have presented a review on SSMs in remote sensing\cite{48}, but their emphasis primarily lies in the comparisons and structural framework of the models across various tasks, such as the specific differences among SSM methods within each task category. In contrast, our paper not only outlines the developmental trajectories of methods across different tasks, covering work from March 2024 to December 2025, as shown in Fig. \ref{fig:development}, but also summarizes eight architectural designs of SSMs. Within each category, we further detail more granular classifications and analyze typical methods along with their strengths. 
    
    In this paper, we present a comprehensive review and analysis of SSMs in remote sensing, systematically tracing their evolution and aiming to provide researchers with a clear understanding of the advancements over the past year. \footnote{We conducted a structured literature search for over 300 RS-SSMs between March 2024 and December 2025. We queried common scholarly databases and digital libraries (e.g., Google Scholar, arXiv, IEEE Xplore, and major publishers' portals) using combinations of keywords including “state space model”,“Mamba", together with “remote sensing”. We included papers that (i) target remote sensing data or tasks, and (ii) use an SSM/Mamba-derived block as a primary modeling component rather than a minor auxiliary module. We removed duplicates across preprint and published versions, and excluded works that only mention SSMs without presenting an SSM-based method. For each included work, we extracted task type, data modality, scanning strategy, architectural design choices, and the key reported metrics.} We also outline potential future research directions, offering insights into how SSMs can continue to evolve to meet the demands of remote sensing applications. By highlighting the strengths and limitations of existing approaches, this review serves as a valuable resource for researchers to develop SSM-based methods for various remote sensing tasks.

    This paper is organized along several key dimensions: 
    \begin{itemize}
    \item 
    Task-Specific adaptations: we analyze how SSMs have been tailored to address the diverse range of tasks in remote sensing, such as classification, object detection, etc., highlighting representative methods. This comprehensive analysis provides valuable guidance for researchers to understand the progress in various tasks, and also helps to identify the strengths of existing methods.
    \item 
    Architectural innovations: we review eight architectural designs of SSMs. By systematically summarizing and analyzing these innovations, our work offers valuable insights into how these innovations specifically tackle the challenges posed by remote sensing data, including dense predictions, high dimensionality, and temporal variations in remote sensing information.
    \item 
    Challenges and opportunities: we summarize the key challenges and opportunities in future research, such as SSM-based remote sensing foundation model, which have not been addressed in previous reviews and are closely linked to the unique characteristics of remote sensing. Our insights can help researchers quickly familiarize themselves with the latest advances and future directions in this field.
    \end{itemize}
    
    The rest of this paper is organized as follows: In Section II, we outline and analyze the development process from the State Space Model to Vision-Mamba. Section III reviews SSMs in remote sensing tasks. Section IV focuses on the key technological advancements of SSM in remote sensing. Section V discusses the current challenges and offers an in-depth analysis of potential research directions for opportunities. Finally, Section VI concludes the paper.

\section{From State Space Model To Vision-Mamba}

In this section, we will introduce the evolution of SSM, transitioning from models designed for sequential modeling to those specifically tailored for visual modeling.
    
\subsection{Preliminary of SSM}

The State Space Model is a mathematical model of a physical system that uses state variables to describe how inputs influence the system’s behavior over time, typically through first-order differential or difference equations. Originating in the 1950s \cite{19}, this approach was applied in fields such as control engineering and system identification. SSM comprises of two key equations: 1) the state equation, which defines the relationship between inputs and previous states through matrix operations, 2) the output equation, which determines how the state matrix is transformed into observable outputs:

\begin{equation}
\begin{split}
    \dot{x}(t) &= Ax(t) + Bu(t) \\
    \dot{y}(t) &= Cx(t) + Du(t)
\end{split}
\end{equation}

where $x(t) \in \mathbb{R}^n$ represents the $n$ state variables, $u(t) \in \mathbb{R}^m$ represents the $m$ state inputs, $y(t) \in \mathbb{R}^p$ represents the $p$ outputs. $A \in \mathbb{R}^{n \times n}$ is the state matrix (controlling the latent state $x$), $B \in \mathbb{R}^{n \times m}$ is the control matrix, $C \in \mathbb{R}^{p \times n}$ is the output matrix, $D \in \mathbb{R}^{p \times m}$ is the command matrix. This structure makes SSM highly versatile, leading to widespread applications in various fields over the past 70 years, including control theory, economic forecasting, and robotics\cite{49,50,51}.

\subsection{SSMs for Sequential Modeling}

With the rapid growth of sequence length and data volume, improving the efficiency of attention-based models has become an important research direction. SSMs provide a complementary alternative to self-attention by enabling linear-time sequence mixing, making it a compelling choice for sequential modeling. In Transformers, the self-attention mechanism results in computational complexity that scales quadratically with the number of tokens, making it challenging to adapt to long-sequence modeling or prediction tasks. In contrast, SSMs benefits from state and output equations that can be interwoven in a manner that mathematically unfolds into a structure akin to one-dimensional convolution operations, resulting in linear complexity. Therefore, in August 2022, Gu \textit{‌et al.} introduced SSMs to sequence processing tasks and developed the S4 model\cite{52} to address the inherent computational complexity of the self-attention mechanism. The S4 model introduces a memory mechanism that incorporates time steps and the HiPPO framework \cite{53}, allowing it to compress all currently observed input signals into a coefficient vector. The discretized SSM can be described in the following form.

\begin{equation}
\begin{split}
h_k&=\bar{A}h_{k-1}+\bar{B}x_k \\
y_k&=Ch_k
\end{split}
\end{equation}

where $h_k\in\mathbb{R}^{n \times l}$ represents the state variable at step $k$. $h_{k-1}\in\mathbb{R}^{n \times l}$ denotes the state variable at step $k-1$, $\bar{A}\in\mathbb{R}^{n\times n}$. $\bar{B}\in\mathbb{R}^{n\times d}$, and $C\in\mathbb{R}^{n\times n}$ are the parameter matrices. $x_k\in\mathbb{R}^{d \times l}$ and $y_k\in\mathbb{R}^{n\times l}$ represent the input and output at step $k$, respectively. $l$, $d$, and $n$ denote the sequence length, input vector dimension, and state size, respectively. 

At the same time, when SSM is stacked in multiple layers, it can be represented in a form similar to one-dimensional convolution, thereby achieving fast inference.

\begin{equation}
\begin{split}
y_{k}&=C\bar{A}^{k}\bar{B}x_{0}+C\bar{A}^{k-1}\bar{B}x_{1}+\cdots+C\bar{A}\bar{B}x_{k-1}+C\bar{B}x_{k}\\
&=\left[C\bar{A}^{k}\bar{B},C\bar{A}^{k-1}\bar{B},\cdots,C\bar{A}\bar{B},C\bar{B}\right]\times\left[x_{0},x_{1},\cdots,x_{k-1},x_{k}\right]
\end{split}
\end{equation}

Early SSMs did not perform as well as self-attention mechanisms in key modalities such as language. Therefore, in December 2023, Gu \textit{‌et al.} introduced two key improvements by proposing the structured SSM (S6), also known as Mamba \cite{1}. At the model level, SSM's parameters are designed to be functions of the input, allowing the model to selectively propagate or forget information throughout the sequence. At the hardware level, it employs a hardware-aware algorithm aimed at minimizing I/O access, thereby achieving further acceleration. In language modeling, S6 achieves an inference speed five times faster than Transformers, while matching the accuracy of Transformers that are twice its size.

\subsection{SSMs for Vision (Vision-Mamba)}

In the realm of vision tasks, where images are highly position-sensitive and requires comprehensive global context, SSMs have garnered widespread attention for their ability to model long-sequence dependencies with linear complexity. As a result, in January 2024, Zhu \textit{‌et al.} sought to introduce SSMs into computer vision applications and developed Vision Mamba\cite{42}. The structure of Vision Mamba is similar to that of Vision Transformer, while the self-attention blocks have been replaced with Mamba blocks. Liu \textit{‌et al.} proposed VMamba \cite{54} during the same period, which introduced 2D-selective scanning. This implementation features various scanning techniques through the 2D Selective Scan (SS2D) block, integrating them with internal Visual State-Space (VSS) blocks. The design of the model structure and scanning strategy for VisionMamba and Vmamba provides insights for subsequent research.

\begin{figure*}
	\centering
	\includegraphics[width=1\linewidth]{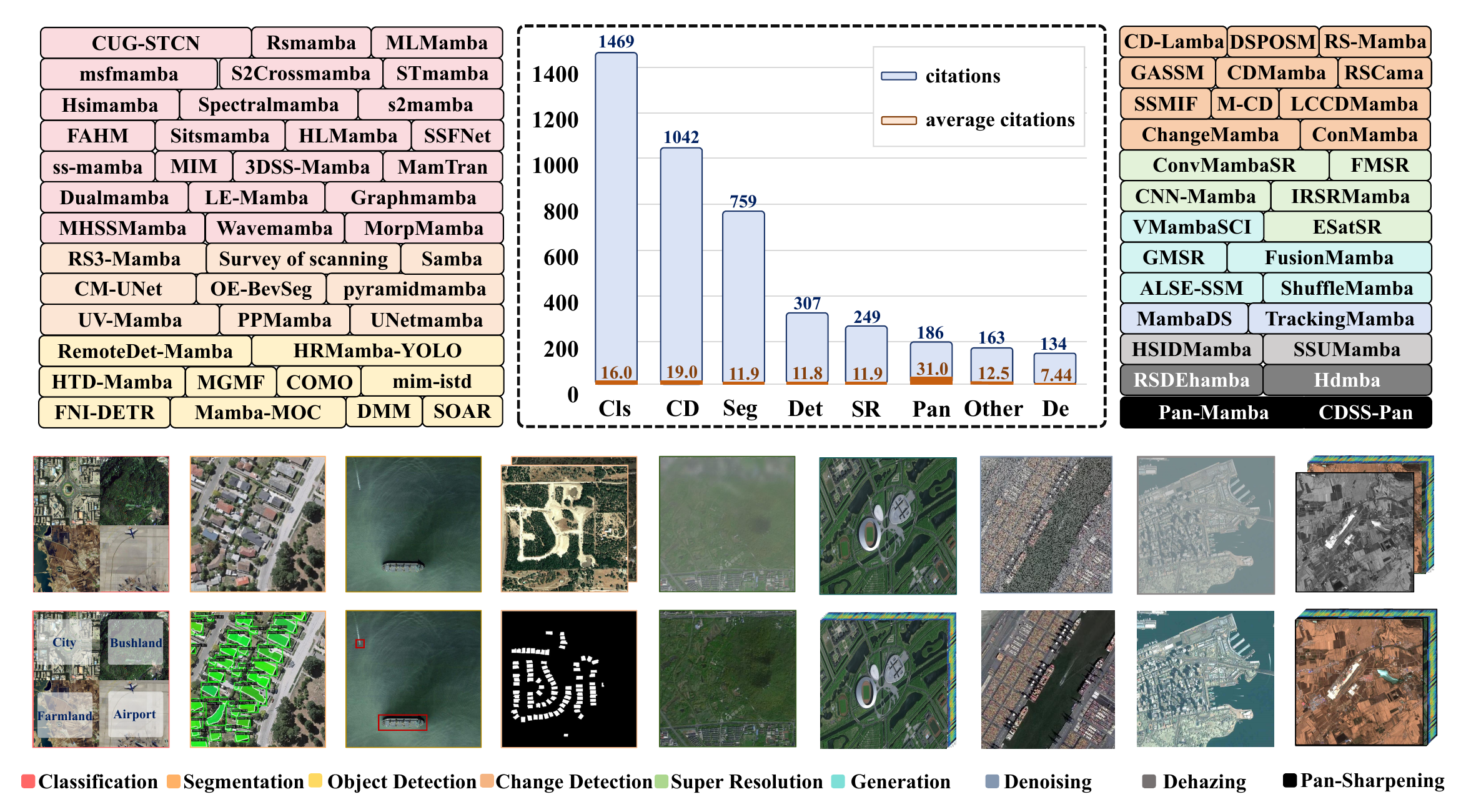}
        \vspace{-4ex}
	\caption{{Statistics on representative remote sensing tasks addressed by SSMs, including Image Classification (Cls), Segmentation (Seg), Object Detection (Det), Change Detection (CD), Super Resolution (SR), Denoising and Dehazing (De), and Pan-sharpening (Pan). We also report the total and average citation counts for papers in each task category, as of December 18, 2025.}}
	\label{fig:hitstory}
\end{figure*}

\begin{table*}[!t]
\renewcommand{\arraystretch}{1.3}
\caption{{Introduction of typical SSMs works in the field of remote sensing. This table shows the categories and highlights, and whether source code are open to public.}}
\label{tab:Comparisons_other_methods_highlight}
\centering
\resizebox{1\linewidth}{!}{
\begin{tabular} {c|c|c|c}
	\toprule
 \multicolumn{1}{c|}{\shortstack{Task}} & {Method} & {Highlight} & {Code}\\
        \midrule
\multirow{17}{*}{\shortstack{Classification}}
&{RSMamba\cite{55}} & {Designing multipath activation scanning to model non-causal data} & {\href{https://github.com/KyanChen/RSMamba}{\textcolor{blue}{link}}} \\
&{HSIMamba\cite{56}} & {Processing remote sensing data bidirectionally} & {\href{https://github.com/Judyxyang/HSImamba}{\textcolor{blue}{link}}} \\
&{SpectralMamba\cite{57}} & {Extracting HSI feature by segmented scanning on frequency bands} &{\href{https://github.com/Judyxyang/HSImamba}{\textcolor{blue}{link}}} \\
&{S$^2$Mamba\cite{58}} & {Capturing spatial contextual relations by patch cross scanning} & {\href{https://github.com/pure-melo/s2mamba}{\textcolor{blue}{link}}} \\
&{SS-Mamba\cite{59}} & {Extracting feature by dual-branch spatial and spectral Mamba} & {\href{https://github.com/mengduanjinghua/spectral-spatial-mamba-for-hsic}{\textcolor{blue}{link}}} \\
&{MIM\cite{60}} & {Designing Mamba-in-mamba architecture for multi-scale feature} & {\href{https://github.com/zhouweilian1904/Mamba-in-Mamba}{\textcolor{blue}{link}}} \\
&{3D SS-Mamba\cite{61}} & {Designing spectral-spatial selective scanning} & {\href{https://github.com/IIP-Team/3DSS-Mamba}{\textcolor{blue}{link}}} \\
&{MamTrans\cite{62}} & {Integrating SSM with attention to improve  accuracy} & {\href{https://github.com/PPPPPsanG/MamTrans}{\textcolor{blue}{link}}} \\
&{DualMamba\cite{63}} & {Designing cross-attention spectral-spatial Mamba} & {-} \\
&{LE-Mamba\cite{64}} & {Fusing SSM with local spectral features} & {-} \\
&{GraphMamba\cite{65}} & {Spatial encoding with Weighted Multi-hop Aggregation (WMA)} & {\href{https://github.com/graphmamba/gmn}{\textcolor{blue}{link}}} \\
&{MHSSMamba\cite{66}} & {Gating mechanisms for spectral-spatial feature enhancement} & {\href{https://github.com/mahmad00/MHSSMamba}{\textcolor{blue}{link}}} \\
&{WaveMamba\cite{67}} & {Combining SSM with wavelet feature linear transformation} & {\href{https://github.com/AlexZou14/Wave-Mamba}{\textcolor{blue}{link}}} \\
&{MorpMamba\cite{68}} & {Designing erosion and dilation based feature processing} & {-} \\
&{MSFMamba\cite{69}} &{Designing multi-scale feature fusion Mamba for multi-modal data} & {\href{https://github.com/oucailab/MSFMamba}{\textcolor{blue}{link}}} \\
&{Enhanced-Mamba\cite{70}} & {Designing multi-scale feature fusion Mamba for multi-modal data} & {-} \\
&{HLMamba\cite{64}} & {Designing a mamba fusion module to fuse multi-modal features} & {-} \\
  \midrule
  \multirow{6}{*}{\shortstack{Object Detection}}
&{MIM-ISTD\cite{71}} & {Designing a mamba-in-mamba architecture for small target} & {\href{https://github.com/txchen-USTC/MiM-ISTD}{\textcolor{blue}{link}}} \\
&{RemoteDet-Mamba\cite{72}} & {Designing a Cross-modal fusion SSM to capture the local features}  & {-} \\
&{Mamba-MOC\cite{73}} & {Designing a cross-scale SSM to capture global and local information } & {-} \\
&{HTD-Mamba\cite{74}} & {Designing a pyramid SSM to capture and fuse spectral-wise features} & {\href{https://github.com/shendb2022/HTD-Mamba}{\textcolor{blue}{link}}} \\
&{DMM\cite{44}} & {Difference-guided multispectral Mamba feature enhancement} & {\href{https://github.com/another-0/dmm}{\textcolor{blue}{link}}} \\
&{MGMF\cite{75}} & {Designing Mask-guided fusion SSM block to enhance target features} & {-} \\
 \midrule
 \multirow{9}{*}{\shortstack{Segmentation}}
&{RS3mamba\cite{76}} & {Designing SSM with integrated spatial and spectral features} & {\href{https://github.com/sstary/SSRS}{\textcolor{blue}{link}}} \\
&{CM-UNet\cite{77}} & {Designing U-shaped Mamba to enhance global-local information fusion} & {\href{https://github.com/XiaoBuL/CM-UNet}{\textcolor{blue}{link}}} \\
&{PyramidMamba\cite{78}} & {Designing multi-scale pyramid fusion to reduce semantic redundancy} & {\href{https://github.com/WangLibo1995/GeoSeg/tree/main}{\textcolor{blue}{link}}} \\
&{OE-BevSeg\cite{79}} & {Designing double surround scanning to perceive the surroundings} & {-} \\
&{RTMamba\cite{80}} & {Designing inverted triangle pyramid pooling feature aggregation} & {-} \\
&{Samba\cite{81}} & {Combining Mamba block with multi-head self-attention} & {\href{https://github.com/samba-team/samba}{\textcolor{blue}{link}}} \\
&{UNetMamba\cite{82}} & {Designing an UNet-SSM decoder for high-resolution image decoding} & {\href{https://github.com/EnzeZhu2001/UNetMamba}{\textcolor{blue}{link}}} \\
&{UV-Mamba\cite{83}} & {Introducing deformable convolution to alleviate SSM memory loss} & {\href{https://github.com/Devin-Egber/UV-Mamba}{\textcolor{blue}{link}}} \\
&{PPMamba\cite{84}} & {Combines a local auxiliary mechanism with an omnidirectional SSM} & {-} \\
 \midrule
 \multirow{5}{*}{\shortstack{Change Detection}}
&{Changemamba\cite{46}} & {Designing a spatio-temporal modeling scanning strategy} & {\href{https://github.com/ChenHongruixuan/MambaCD}{\textcolor{blue}{link}}} \\
&{RSM\cite{45}} & {Designing a omnidirectional selective scanning module} & {\textcolor{blue}{link}} \\
&{RSCama\cite{85}} & {Designing a SSM with integrated temporal features} & {\href{https://github.com/Chen-Yang-Liu/RSCaMa}{\textcolor{blue}{link}}} \\
&{CDMamba\cite{27}} & {Designing an adaptive global local guided fusion block} & {\href{https://github.com/zmoka-zht/CDMamba}{\textcolor{blue}{link}}} \\
&{M-CD\cite{47}} & {Designing a multi-scale score checking module} & {\href{https://github.com/JayParanjape/M-CD}{\textcolor{blue}{link}}} \\
 \midrule
\multirow{4}{*}{\shortstack{Super Resolution}}
&{FreMamba\cite{86}} & {Designing SSM block with internal embedded Fourier transform}  & {\href{https://github.com/XY-boy/FreMamba}{\textcolor{blue}{link}}} \\
&{IRSRMamba\cite{87}} & {Designing SSM block with internal embedded Wavelet transform} & {\href{https://github.com/yongsongH/IRSRMamba}{\textcolor{blue}{link}}} \\
&{Esatsr\cite{88}} & {Enhancing the global-local feature of SSM} & {-} \\
&{MambaFormerSR\cite{89}} & {Combining attention mechanism with mamba} & {-} \\

 \midrule
 \multirow{2}{*}{\shortstack{Denoising}}
&{HSIDMamba\cite{90}} & {Designing 8-directions scanning strategy} &  {-} \\
&{SSUMamba\cite{91}} & {Designing spatial-spectral continuous scanning strategy} &  {\href{https://github.com/lronkitty/SSUMamba}{\textcolor{blue}{link}}} \\
 \midrule
 \multirow{2}{*}{\shortstack{Recovery}}
&{Rsdehamba\cite{92}} & {Designing a U-Net structure and mambaformer variant} & {-} \\
&{Hdmba\cite{93}} & {Designing a novel window SSM to capture local and global correlations} & {\href{https://github.com/RsAI-lab/HDMba}{\textcolor{blue}{link}}} \\
 \midrule
 \multirow{1}{*}{\shortstack{Pan-sharpening}}
&{Pan-Mamba\cite{43}} & {Designing channel exchange and cross-modal scanning strategy} &  {\href{https://github.com/alexhe101/Pan-Mamba}{\textcolor{blue}{link}}} \\
 \midrule
 \multirow{1}{*}{\shortstack{Meteorological Downscaling}}
&{MambaDS\cite{94}} & {Using SSM to enhance the utilization of multivariate correlations} & {-} \\
	\bottomrule
\end{tabular}
}
\end{table*}

\section{SSMs in Remote Sensing Tasks}

In this section, we review the application of SSMs across various remote sensing tasks, including classification, detection, change detection, semantic segmentation, denoising, super-resolution, dehazing, and pan-sharpening. Fig. \ref{fig:hitstory} illustrates the typical SSMs methods and the total and average citation counts for papers in each task category. Table \ref{tab:Comparisons_other_methods_highlight} shows the categories and highlights, and whether source code are open to public. It demonstrates the popularity of SSM research in the field of remote sensing.

\subsection{SSMs for Remote Sensing Image Classification}

The research on SSMs in remote sensing image classification broadly falls into two main categories.

The first category focuses on designing different scanning strategies to balance computational efficiency with the ability to capture global context \cite{57,58,66,69,102,103}. For example, RSMamba\cite{102} is a typical example of the first approach, which tackled the inherent limitations of SSMs in spatial sensitivity and unidirectional sequence modeling by innovatively implementing multi-directional scanning strategies (forward, reverse, and random). This approach significantly enhanced spatial-context awareness in dynamic remote sensing scenes while establishing a critical paradigm for subsequent SSM-based architectures. Furthermore, RSM-SS \cite{45} and other methods\cite{57,58,66,69,103} have investigated a range of scanning strategies to better capture context.

The second category focuses on designing micro and macrostructure variations that combine SSMs with methods such as CNNs, Transformers, and Graph Neural Networks (GNNs), allowing the model to leverage the unique strengths of each framework \cite{56,59,62,63,67,104,105,106}. For example, HSIMamba \cite{56} is a typical representative of this category; it synergistically integrates the sequential processing of spectral bands by SSMs with the spatial attention mechanisms of CNNs. This hybrid design efficiently models both spectral and spatial information in hyperspectral images. Furthermore, this approach has been adopted by methods such as SpectralMamba and 3DSS-Mamba, which apply SSMs to 3D data while retaining their linear computational scalability \cite{57,61}.

Methods primarily based on scanning strategies can alleviate the bias of unidirectional scanning through multi-directional scanning, making better use of image information. These methods are more suitable for sequential data such as HSI. However, they are relatively less sensitive to local texture details. In contrast, methods focused on structural mixing can effectively compensate for modeling local details and key information, but they also increase the computational burden of the model.

Table \ref{tab:Comparisons_other_methods_houston2013} and Table \ref{tab:Comparisons_other_methods_indian} present comparisons of hyperspectral image classification models on the Houston2013 dataset\cite{95} and the Indian dataset\cite{107}, respectively. In our table, the data is sourced from the original papers and their code repositories, which can be accessed through the GitHub repository we maintain. The shaded cells indicate the top three performing models, with darker colors representing higher rankings. 

Table \ref{tab:Comparisons_other_methods_houston2013} showcases the results on the Houston2013 dataset\cite{95}, which consists of 15 classes. Among these, Classes 1-4 represent different vegetations, while Classes 5-6 represent natural environments (soil and water). Classes 7-8 represent extensive building coverage, which requires contextual scene understanding and a larger receptive field for differentiation. Classes 9-11 correspond to road systems, both of which are geometrically narrow structures. Classes 12-15 represent targets with distinct optical features but are relatively small in size. Based on the class-wise performance scores, some notable characteristics of RS-SSMs can be identified: (i) For Classes 1-4, which possess similar spectral features, 3DSSMamba\cite{61} and HSIMamba\cite{56} employ spectral-spatial joint scanning models that effectively reduce misclassifications. (ii) Classes 12-15 require the model to have strong capabilities for small target learning. Methods such as DualMamba\cite{63}, which leverage local feature fusion, achieve notable performance improvements in these scenarios.

Table \ref{tab:Comparisons_other_methods_indian} illustrates the results on the Indian dataset\cite{107}, which includes 16 classes. This dataset features crops at different growth stages and has a limited total number of samples. Traditional local convolution-based models or unidirectional spectral modeling approaches tend to struggle with inter-class confusion in such cases. However, SSM's multi-directional scanning effectively mitigates this issue. Additionally, the dataset is highly imbalanced in terms of class distribution; for instance, Classes 1, 7, and 9 each contain no more than 10 images. Nevertheless, SSM exhibits minimal performance discrepancies across different classes, highlighting its ability to learn from a small number of samples. This characteristic warrants further exploration in future studies.

\begin{table*}[!t]  
\renewcommand{\arraystretch}{1.3}
\caption{{Comparison results of classical methods and previous SSMs methods on remote sensing classification dataset Houston2013\cite{95}. The comparison metrics include: accuracy for each category (C1 to C15), overall accuracy (OA), average accuracy (AA), and kappa coefficient ($\kappa$).}}
\label{tab:Comparisons_other_methods_houston2013}
\centering

\resizebox{\textwidth}{!}{
\begin{tabular}{c|c c c c c c c c c c c c c c c c c c}

	\toprule

 \multicolumn{1}{c|}{\shortstack{Method}} & { C1} & C2 & { 
 C3} & C4 & C5 & C6 & C7 & C8 & C9 & C10 & C11 & C12 & C13 & C14 & C15 & {OA(\%)} & {AA(\%)} & {$\kappa$}\\
        \midrule

\multicolumn{1}{c|}{1d CNN\cite{96}} &87.3&98.21&\cellcolor{darkgray}100.0&92.99&97.35&95.10&77.33&51.38&27.95&90.83&79.32&76.56&69.47&99.19&98.10 &80.04&82.74&78.35\\
\multicolumn{1}{c|}{2d CNN\cite{97}} &85.1&\cellcolor{lightgray}99.91&77.23&97.73&99.53&92.31&92.16&79.39&86.31&43.73&87.00&66.28&90.18&90.69&77.80&83.72&84.35&82.31\\

\multicolumn{1}{c|}{RNN\cite{98}} &	82.3&	94.27&	\cellcolor{lightgray}99.60&	97.54&	93.28&	95.10&	83.77&	56.03&	72.14&	84.17&	82.83&	70.61&	69.12&	98.79&	95.98&	83.23&	85.04&	81.83\\

\multicolumn{1}{c|}{mini GCN\cite{99}} &	98.4&	99.11&	\cellcolor{lightgray}99.60&	96.68&	97.73&	95.10&	76.77&	68.09&	53.92&	77.41&	84.91&	77.23&	50.88&	98.38&	98.52&	81.71&	83.09&	80.18\\

\multicolumn{1}{c|}{ViT\cite{100}} &	82.6&	92.82&	\cellcolor{gray}99.80&	99.24&	97.73&	95.10&	76.77&	55.65&	67.42&	68.05&	82.35&	58.50&	60.00&	98.79&	98.73&	80.41&	82.50&	78.76\\
\multicolumn{1}{c|}{Spectralformer\cite{101}} &	81.9
&	\cellcolor{darkgray}100.0&	95.25&	96.12&	99.53&	94.41&	83.12&	76.73&	79.32&	78.86&	88.71&	87.32&	72.63&	\cellcolor{darkgray}100.0&	\cellcolor{gray}99.79&	86.14&	87.48&	84.97\\
  \midrule
\multicolumn{1}{c|}{HSIMamba\cite{56}} &	\cellcolor{darkgray}99.8
&	\cellcolor{lightgray}99.34&	\cellcolor{darkgray}100.0&	99.34&	\cellcolor{darkgray}100.0&	\cellcolor{darkgray}100.0&	\cellcolor{lightgray}94.96&	95.06&	\cellcolor{lightgray}93.86&	91.60&	97.06&	97.69&	\cellcolor{gray}99.65&	\cellcolor{darkgray}100.0&	\cellcolor{darkgray}100.0&	\cellcolor{lightgray}97.29&	\cellcolor{lightgray}97.89&	\cellcolor{lightgray}97.06\\
\multicolumn{1}{c|}{SpectralMamba\cite{57}} &95.9&	98.87&	98.61&	99.05&	\cellcolor{gray}99.91&	94.41&	87.03&	64.86&	80.64&	98.17&	80.36&	81.56&	79.65&	\cellcolor{darkgray}100.0&	98.52&	89.52&	90.50&	88.64\\
\multicolumn{1}{c|}{S$^2$Mamba\cite{58}} &	83.1&	\cellcolor{darkgray}100.0&	\cellcolor{lightgray}99.60&	98.20&	\cellcolor{darkgray}100.0&	95.80&	89.37&	88.60&	92.45&	92.57&	91.56&	90.97&	89.12&	\cellcolor{darkgray}100.0&	\cellcolor{darkgray}100.0&	93.36&	94.09&	92.79\\
\multicolumn{1}{c|}{SSMamba\cite{59}} &	92.9&	95.99&	\cellcolor{darkgray}100.0&	99.17&	98.29&	\cellcolor{lightgray}96.39&	91.60&	83.33&	92.05&	98.05&	92.48&	91.24&	92.87&	\cellcolor{darkgray}100.0&	\cellcolor{darkgray}100.0&	94.30&	94.96&	93.84\\
\multicolumn{1}{c|}{3DSS-Mamba \cite{61}} &	\cellcolor{gray}99.0&	\cellcolor{gray}99.56&	\cellcolor{darkgray}100.0&	97.68&	\cellcolor{lightgray}99.91&	\cellcolor{gray}98.29&	\cellcolor{gray}97.55&	96.16&	\cellcolor{gray}96.27&	99.55&	\cellcolor{lightgray}98.20&	\cellcolor{lightgray}98.38&	\cellcolor{lightgray}96.68&	\cellcolor{lightgray}99.48&	\cellcolor{darkgray}100.0&	\cellcolor{gray}98.37&	\cellcolor{gray}98.44&	\cellcolor{gray}98.24\\
\multicolumn{1}{c|}{DualMamba \cite{63}} &	\cellcolor{lightgray}98.8&	\cellcolor{darkgray}100.0&	\cellcolor{darkgray}100.0&	98.75&	\cellcolor{darkgray}100.0&	\cellcolor{darkgray}100.0&	\cellcolor{darkgray}98.55&	\cellcolor{darkgray}99.60&	\cellcolor{darkgray}98.22&	\cellcolor{darkgray}100.0&	\cellcolor{gray}99.73&	\cellcolor{darkgray}99.63&	\cellcolor{darkgray}100.0&	\cellcolor{darkgray}100.0&	\cellcolor{darkgray}100.0&	\cellcolor{darkgray}99.57&	\cellcolor{darkgray}99.67&	\cellcolor{darkgray}99.51\\
\multicolumn{1}{c|}{GraphMamba\cite{65}} &	96.6&	98.86&	\cellcolor{darkgray}100.0&	\cellcolor{darkgray}100.0&	\cellcolor{darkgray}100.0&	\cellcolor{darkgray}100.0&	94.40&	83.52&	83.36&	\cellcolor{lightgray}99.70&	\cellcolor{darkgray}99.92&	93.57&	93.47&	\cellcolor{darkgray}100.0&	\cellcolor{darkgray}100.0&	95.62&	96.23&	95.27\\
\multicolumn{1}{c|}{MHSS-Mamba\cite{66}} &	97.6&	99.84&	99.13&	99.03&	99.83&	95.67&	93.53&	\cellcolor{lightgray}96.94&	92.17&	\cellcolor{gray}99.83&	95.95&	94.97&	89.74&	97.66&	\cellcolor{lightgray}99.09&	96.92&	96.73&	96.67\\
\multicolumn{1}{c|}{WaveMamba\cite{67}} &	97.9&	98.72&	\cellcolor{darkgray}100.0&	98.39&	99.67&	93.82&	87.85&	\cellcolor{gray}97.42&	91.69&	98.20&	96.27&	\cellcolor{gray}98.70&	82.90&	\cellcolor{gray}99.53&	\cellcolor{darkgray}100.0&	96.39&	96.07&	96.1\\
\multicolumn{1}{c|}{MamTrans\cite{62}} &	82.7&	\cellcolor{darkgray}100.0&	98.61&	\cellcolor{gray}99.81&	\cellcolor{darkgray}100.0&	95.80&	87.59&	95.44&	83.95&	91.99&	97.53&	90.01&	84.91&	\cellcolor{darkgray}100.0&	\cellcolor{darkgray}100.0&	93.41&	93.89&	92.84\\

	\bottomrule
\end{tabular}}
\end{table*}

\begin{table*}[!t]
\renewcommand{\arraystretch}{1.3}
\caption{{Comparison results of classical methods and previous SSMs methods on remote sensing classification dataset Indian\cite{107}. The comparison metrics include: accuracy for each category (C1 to C16), overall accuracy (OA), average accuracy (AA), and kappa coefficient ($\kappa$).}}
\label{tab:Comparisons_other_methods_indian}
\centering

\resizebox{\linewidth}{!}{

\begin{tabular}{c|c c c c c c c c c c c c c c c c c c c}
	\toprule

 \multicolumn{1}{c|}{\shortstack{Method}} & C1 & C2 &C3 & C4 & C5 & C6& C7 & C8 &C9 & C10 & C11 & C12 & C13 & C14 &C15 & C16 & {OA(\%)} & {AA(\%)} & {$\kappa$}\\
        \midrule

\multicolumn{1}{c|}{1d CNN\cite{96}} &	47.83&	42.35&	60.87&	89.49&	97.04&	59.69&	64.81&	48.68&	44.33&	\cellcolor{lightgray}96.30&	74.28&	15.45&	91.11&	33.33&	\cellcolor{darkgray}100.0&	80.00&	70.43&	79.60&	66.42\\
\multicolumn{1}{c|}{2d CNN\cite{97}} &	65.90&	76.66&	72.39&	93.96&	97.27&	77.23&	58.85&	57.03&	72.87&	93.44&	\cellcolor{darkgray}100.0&	88.18&	\cellcolor{darkgray}100.0&	84.62&	\cellcolor{darkgray}100.0&	\cellcolor{darkgray}100.0&	75.89&	86.64&	72.81\\
\multicolumn{1}{c|}{RNN\cite{98}} &	69.00&	58.93&	77.17&	82.33&	97.07&	69.06&	53.56&	63.06&	65.07&	95.06&	88.67&	50.00&	\cellcolor{lightgray}97.78&	66.67&	81.82&	\cellcolor{darkgray}100.0&	70.66&	76.37&	66.73\\
\multicolumn{1}{c|}{mini GCN\cite{99}} &	72.54&	55.99&	92.93&	92.62&	98.63&	64.71&	68.78&	63.38&	69.33&	98.77&	87.78&	50.00&	\cellcolor{darkgray}100.0&	48.72&	72.73&	80.00&	75.11&	78.03&	71.64\\
\multicolumn{1}{c|}{ViT\cite{100}} &	53.25&	66.20&	88.60&	89.71&	89.98&	72.22&	66.00&	57.09&	57.09&	97.53&	87.62&	63.94&	95.56&	79.49&	90.91&	80.00&	71.86&	78.97&	68.04\\
\multicolumn{1}{c|}{Spectralformer\cite{101}} &	70.52&	81.39&	91.30&	95.53&	\cellcolor{lightgray}99.32&	81.81&	75.48&	73.76&	73.76&	98.77&	93.17&	78.48&	\cellcolor{darkgray}100.0&	79.49&	\cellcolor{darkgray}100.0&	\cellcolor{darkgray}100.0&	81.76&	87.81&	79.19\\
  \midrule
\multicolumn{1}{c|}{HSIMamba\cite{56}} &	\cellcolor{darkgray}100.0&	80.36&	79.89&	95.97&	90.39&	\cellcolor{gray}99.99&	\cellcolor{darkgray}100.0&	\cellcolor{gray}99.79&	\cellcolor{darkgray}100.0&	83.95&	51.77&	85.45&	\cellcolor{darkgray}100.0&	79.49&	90.91&	\cellcolor{darkgray}100.0&	89.92&	89.82&	88.57\\
\multicolumn{1}{c|}{S$^2$Mamba\cite{58}} &	\cellcolor{gray}94.44&	\cellcolor{darkgray}100.0&	\cellcolor{darkgray}100.0&	\cellcolor{gray}98.43&	\cellcolor{darkgray}100.0&	\cellcolor{darkgray}100.0&	\cellcolor{gray}98.47&	98.10&	\cellcolor{lightgray}95.04&	\cellcolor{darkgray}100.0&	97.67&	\cellcolor{darkgray}100.0&	\cellcolor{darkgray}100.0&	\cellcolor{darkgray}100.0&	\cellcolor{darkgray}100.0&	\cellcolor{darkgray}100.0&	\cellcolor{darkgray}97.92&	\cellcolor{darkgray}98.88&	\cellcolor{darkgray}97.61\\
\multicolumn{1}{c|}{SSMamba\cite{59}} &	\cellcolor{darkgray}100.0&	80.30&	88.54&	\cellcolor{darkgray}99.49&	93.69&	98.34&	\cellcolor{darkgray}100.0&	\cellcolor{darkgray}100.0&	\cellcolor{darkgray}100.0&	89.61&	89.51&	\cellcolor{gray}93.32&	\cellcolor{gray}99.46&	\cellcolor{lightgray}98.63&	\cellcolor{lightgray}96.67&	\cellcolor{gray}99.73&	91.59&	95.46&	90.42\\
\multicolumn{1}{c|}{3DSS-Mamba\cite{61}} &	78.05&	\cellcolor{lightgray}90.27&	93.04&	93.43&	97.70&	97.26&	\cellcolor{lightgray}96.00&	\cellcolor{lightgray}99.30&	38.89&	\cellcolor{gray}96.80&	\cellcolor{gray}98.91&	91.39&	95.68&	\cellcolor{gray}99.65&	94.52&	\cellcolor{lightgray}84.52&	95.82&	90.83&	95.23\\
\multicolumn{1}{c|}{GraphMamba\cite{65}} &	\cellcolor{darkgray}100.0&	\cellcolor{gray}94.25&	\cellcolor{lightgray}94.41&	\cellcolor{lightgray}98.39&	94.63&	\cellcolor{lightgray}99.71&	\cellcolor{darkgray}100.0&	\cellcolor{darkgray}100.0&	\cellcolor{darkgray}100.0&	95.33&	95.30&	\cellcolor{lightgray}93.31&	\cellcolor{darkgray}100.0&	\cellcolor{darkgray}100.0&	\cellcolor{gray}99.91&	\cellcolor{darkgray}100.0&	\cellcolor{lightgray}96.43&	\cellcolor{lightgray}97.83&	\cellcolor{lightgray}95.91\\
\multicolumn{1}{c|}{MamTrans\cite{62}} &	\cellcolor{lightgray}94.00&	\cellcolor{darkgray}100.0&	\cellcolor{gray}99.45&	\cellcolor{gray}98.43&	\cellcolor{gray}99.86&	\cellcolor{darkgray}100.0&	91.83&	97.06&	\cellcolor{gray}95.92&	\cellcolor{darkgray}100.0&	\cellcolor{lightgray}98.24&	\cellcolor{darkgray}100.0&	\cellcolor{darkgray}100.0&	\cellcolor{darkgray}100.0&	\cellcolor{darkgray}100.0&	\cellcolor{darkgray}100.0&	\cellcolor{gray}97.07&	\cellcolor{gray}98.42&	\cellcolor{gray}96.64\\

	\bottomrule
\end{tabular}}

\end{table*}

\subsection{SSMs for Remote Sensing Image Segmentation}

Semantic segmentation is a dense image processing task that requires models to simultaneously understand both global semantics and local details. There are two main approaches to incorporating SSMs in semantic segmentation. 

The first category focuses on leveraging SSMs and CNN to separately extract global and local information, which are then fused together \cite{76,84,108}. For example, RS3Mamba\cite{76} is a typical example of the first approach, which was developed shortly after  RSMamba\cite{102} and Pan-Mamba \cite{43}. RS3Mamba employs a residual network as its main branch while utilizing an auxiliary encoder composed of Mamba to enhance global dependencies. RS3Mamba demonstrates superior performance across various datasets, producing less noise and achieving smoother boundaries in segmentation results.

The second category \cite{77,78,81} focuses on integrating SSMs with network architectures such as U-Net \cite{114}. For example, Liu \textit{‌et al.} developed CM-UNet\cite{77}, which also follows a typical U-shaped architecture but replaces the traditional skip connections with SSM blocks. The SSM blocks retains a gated structure, and integrate multiscale convolutional kernels (3x3, 5x5, and 7x7) to process combined features. The spatial features are then aggregated through mean and max pooling.

The first category balances global semantics and local boundaries, resulting in smoother boundaries. However, the additional branches increase the complexity of feature alignment and fusion, leading to increased inference latency. The second category employs a mature multi-scale framework, but most current implementations only replace parts of SSM locally. There is a need for further research to fully utilize SSM for modeling multi-scale information.

Table \ref{tab:Comparisons_other_methods_lovada} presents comparisons of these models on the Lovada dataset\cite{109}. UNetMamba and Samba achieved better results, and their commonality lies in the direct construction of multi-scale features. SSM itself does not incorporate multi-scale information, and this construction method compensates for the limitations of SSM. 

\begin{table*}[!t]
\renewcommand{\arraystretch}{1.3}
\caption{{Comparison results of classical methods and previous SSMs methods on remote sensing semantic segmentation dataset LoveDA\cite{109}.}}
\label{tab:Comparisons_other_methods_lovada}
\centering

\resizebox{0.7\linewidth}{!}{

\begin{tabular}{m{1.5cm}|cccccccc}
	\toprule
 \multicolumn{1}{c|}{\shortstack{Method}} & background&building&road&		water&barren&forest&agriculture&MIoU\\
        \midrule
\multicolumn{1}{c|}	{	PidNet	\cite{110}}	&	44.15	&	54.72	&	\cellcolor{gray}56.54	&	76.81	&	17.49	&	\cellcolor{darkgray}47.13	&	57.54	&	\cellcolor{darkgray}50.63	\\
\multicolumn{1}{c|}	{	Manet	\cite{111}}	&	43.29	&	55.35	&	\cellcolor{lightgray}56.2	&	74.18	&	14.28	&	42.85	&	58.62	&	49.26	\\
\multicolumn{1}{c|}	{	TransUNet\cite{112}	}	&	43.00	&	56.10	&	53.70	&	\cellcolor{lightgray}78.00	&	9.30	&	44.07	&	57.85	&	48.31	\\
 \midrule
\multicolumn{1}{c|}{	RS3mamba	\cite{76}}	&	41.60	&	\cellcolor{gray}58.23	&	54.03	&	77.34	&	17.97	&	43.81	&	\cellcolor{gray}61.37	&	\cellcolor{gray}50.62\\
\multicolumn{1}{c|}{	UNetMamba	\cite{82}}	&	\cellcolor{gray}47.08	&	\cellcolor{darkgray}59.16	&	\cellcolor{darkgray}56.74	&	\cellcolor{darkgray}81.37	&	18.15	&	\cellcolor{gray}46.61	&	\cellcolor{darkgray}64.31	&	53.35\\
\multicolumn{1}{c|}{	Samba	\cite{81}}	&	\cellcolor{darkgray}51.71	&	\cellcolor{lightgray}57.85	&	49.86	&	59.85	&	\cellcolor{darkgray}21.93	&	41.00	&	47.56	&	47.11\\
\multicolumn{1}{c|}{	RTmamba-s	\cite{80}}	&	42.40	&	51.06	&	50.45	&	76.09	&	18.25	&	43.81	&	\cellcolor{lightgray}58.51	&	48.65\\
\multicolumn{1}{c|}{	RTmamba-b	\cite{80}}	&	\cellcolor{lightgray}44.72	&	56.43	&	52.49	&	\cellcolor{gray}79.90	&	\cellcolor{gray}20.96	&	\cellcolor{lightgray}44.46	&	55.06	&	\cellcolor{lightgray}50.57\\
\multicolumn{1}{c|}{	SctNet-s	\cite{113}}	&	38.17	&	46.06	&	51.06	&	75.70	&	\cellcolor{lightgray}18.50	&	42.19	&	54.44	&	47.02\\

\multicolumn{1}{c|}{	SctNet-b	\cite{113}}	&	43.37	&	52.17	&	50.51	&	75.71	&	18.15	&	43.21	&	52.97	&	47.73\\

	\bottomrule
\end{tabular}
}
\end{table*}

\begin{table*}[!t]
\renewcommand{\arraystretch}{1.3}
\caption{{Comparison results of classical methods and previous SSMs methods on remote sensing change detection dataset WHU\cite{116}, Levir-CD\cite{117}, and Levir-CD+\cite{117}.}}
\label{tab:Comparisons_other_methods_whu}
\centering

\resizebox{0.7\linewidth}{!}{
\begin{tabular}{c|c c c|c c c|c c c}

	\toprule

\multirow{2}{*}{\shortstack{Method}}
& \multicolumn{3}{c|}{WHU\cite{116}} & \multicolumn{3}{c|}{LEVIR\cite{117}} & \multicolumn{3}{c}{LEVIR+\cite{117}} \\ 

& F1 & IoU & OA(\%) & F1 & IoU & OA(\%) & F1 & IoU & OA(\%) \\ 
        \midrule
\multicolumn{1}{c|}	{	FC-EF	\cite{118}}	&	91.36	&	84.10 	&	99.32	&	88.90 	&	80.03	&	98.89	&	76.41	&	61.82	&	98.08	\\
\multicolumn{1}{c|}	{	IFNet	\cite{119}}	&	89.81	&	81.50 	&	99.16	&	89.69	&	81.31	&	98.96	&	79.23	&	65.61	&	98.34	\\
\multicolumn{1}{c|}	{	SwinsUnet	\cite{120}}	&	89.93	&	81.71	&	99.22	&	87.77	&	78.21	&	98.77	&	78.31	&	64.35	&	98.22	\\
\multicolumn{1}{c|}	{	ChangeFormer	\cite{121}}	&	90.30 	&	82.32	&	99.26	&	88.83	&	79.90 	&	98.88	&	77.54	&	63.31	&	98.16	\\
\midrule
\multicolumn{1}{c|}	{	RSM	\cite{45}}	&	\cellcolor{lightgray}92.79	&	\cellcolor{lightgray}86.55	&	\cellcolor{lightgray}99.44	&	89.77	&	81.44	&	98.97	&	\cellcolor{lightgray}80.91	&	\cellcolor{lightgray}67.95	&	\cellcolor{lightgray}98.42	\\
\multicolumn{1}{c|}	{	ChangeMamba	\cite{46}}	&	92.55	&	86.13	&	99.42	&	\cellcolor{lightgray}90.16	&	\cellcolor{lightgray}82.09	&	\cellcolor{lightgray}99.01	&	80.77	&	67.74	&	98.41	\\
\multicolumn{1}{c|}	{	CDMamba	\cite{27}}	&	\cellcolor{gray}93.76	&	\cellcolor{gray}88.26	&	\cellcolor{gray}99.51	&	\cellcolor{gray}90.75	&	\cellcolor{gray}83.07	&	\cellcolor{gray}99.06	&	\cellcolor{gray}83.01	&	\cellcolor{gray}70.95	&	\cellcolor{gray}98.65	\\
\multicolumn{1}{c|}	{	M-CD	\cite{47}}	&	\cellcolor{darkgray}95.3	&	\cellcolor{darkgray}91.1	&	\cellcolor{darkgray}99.6	&	\cellcolor{darkgray}92.1	&	\cellcolor{darkgray}85.0 	&	\cellcolor{darkgray}99.2	&	-	&	-	&	-	\\
	\bottomrule
\end{tabular}
}
\end{table*}

\subsection{SSMs for Remote Sensing Object Detection}

Current research on SSMs in object detection broadly falls into two main categories, reflecting different strategies to handle the unique challenges of remote sensing data.

The first approach focuses on leveraging SSMs to enable long-sequence modeling, making full use of both local and distant contextual information around each pixel\cite{74,115}. For example, HTD-Mamba\cite{74} is a typical example of the first approach proposed by Shen \textit{‌et al.}. For each pixel's surrounding area, HTD-Mamba \cite{74} utilizes a pyramid state space model, performs multi-scale downsampling of the input SSMs token sequence. Following the extraction of features through the SSM, it re-fuses these features via upsampling. This approach effectively leverages both local and remote spatial information surrounding each pixel, highlighting the competitiveness and potential of SSMs in object detection tasks. 

The second approach emphasizes combining SSMs with multimodal information, designing specialized scanning methods to effectively fuse data from different sensors, such as hyperspectral and multispectral images\cite{44,75}. For example, DMM \cite{44} and MGMF \cite{75} are typical examples of the second approach, which further explores the capabilities of SSMs in object detection across multiple modalities. They designed various model variants of SSMs and investigated the effectiveness of multimodal feature fusion.

The first approach is more favorable for small and dense targets, achieving a balance between the receptive field range and efficiency. The second approach leverages SSM's selective gating to suppress cross-modal redundant information. However, this may lead to fluctuating performance in cases of modality loss or uneven quality.

\subsection{SSMs for Remote Sensing Change Detection}

In SSM-based change detection (CD), a common approach is to arrange multi-temporal data in different sequences and use SSMs to capture the surface changes between multi-temporal remote sensing images.

The ChangeMamba \cite{46} proposed by Chen \textit{‌et al.} is the typical SSMs model in remote sensing CD tasks, addressing tasks including binary change detection, building damage assessment, and semantic change detection. Three modeling approaches were utilized: 1) spatio-temporal sequential modeling, which unfolds the tokens of two temporal phases of data and then arranges them in temporal order; 2) spatio-temporal cross modeling, where the tokens of the two temporal phases are processed in a cross-ordered fashion; and 3) spatial-temporal parallel modeling, which concatenates the tokens of the two temporal phases along the channel dimension. This exploration has established notable methodologies for subsequent change detection efforts. Unlike ChangeMamba \cite{46}, which utilizes State Space Models (SSM) for feature extraction, subsequent works such as M-CD \cite{122} and RSCama \cite{85} investigate multi-temporal feature fusion through SSMs by defining scanning orders for sequential data to capture spatiotemporal relationships. These methods provide strong evidence for understanding the complex relationships between images at different times. However, it remains to be explored which modeling approach is optimal for specific change detection tasks and diverse types of land cover distribution characteristics.

Table \ref{tab:Comparisons_other_methods_whu} presents comparisons of these models on datasets WHU\cite{116}, Levir-CD\cite{117}, and Levir-CD+\cite{117}. CDmamba and M-CD both focus on modifying the internal structure of SSM by introducing additional branches and utilizing SSM for the fusion of multi-temporal features, thereby maximizing SSM's capability to capture key differential information.

\subsection{SSMs for Remote Sensing Image Super Resolution}
Research on applying SSM to image super-resolution tends to utilize Fourier transforms and other frequency-domain transformation methods, integrating SSMs blocks to achieve SSMs scanning at the frequency domain level \cite{86,87}.

FMSR\cite{86} is a typical representative and marks the first attempt at applying SSMs in remote sensing image super-resolution. FMSR  employs a frequency selection module based on two-dimensional fast Fourier transforms to process features, while also incorporating a VSSM branch to extract remote features. To effectively handle local features, a hybrid gate module (HGM) is introduced. The subsequent IRSRMamba\cite{87} follows a similar framework, utilizing wavelet transforms.

These approaches utilize frequency domain methods such as Fourier transform and the wavelet transform, integrated with SSM blocks, and can capture the global structural information of images. SSM can selectively model this information, combined with CNNs to process local details, achieving excellent reconstruction results. However, these approaches face the challenge of effectively fusing features extracted from the frequency domain with those from the spatial domain. Suboptimal fusion strategies may lead to information conflict or insufficient interaction, ultimately hampering the overall reconstruction performance.

\subsection{SSMs for Remote Sensing Image Denoising and Dehazing}

Currently, the exploration of SSMs in denoising and dehazing is still in the exploratory stage, primarily focusing on denoising and dehazing.

HSIDMamba \cite{90} addresses hyperspectral image (HSI) denoising through State Space Model scanning strategies. HSIDMamba \cite{90} incorporates eight different scanning starting points and directions, allowing the extracted features to be processed by the State Space Model in multiple orientations. Additionally, HSIDMamba \cite{90}  incorporates a spectral attention-gating structure to filter out redundant features and enhance denoising performance through average pooling and CNN activation.

Remote Sensing Image Dehazing focuses on removing uneven and physically irregular haze factors to achieve high-quality image restoration. Zhou \textit{‌et al.} explored the feasibility of SSMs for dehazing tasks \cite{92}. RSDeHamba \cite{92} has a relatively simple, U-shaped architecture, akin to FusionMamba \cite{123}. The network alternates between vision dehamba blocks and up-sampling techniques during the reconstruction process. The vision dehamba block is a variant of Mambaformer \cite{89} and incorporates multi-directional State Space Modeling, convolution operations, and channel attention mechanisms.

These studies provide new perspectives for tasks traditionally dominated by CNNs, as the long-range modeling capability of SSM helps address non-uniform noise or haze. Moreover, compared to other tasks, related research is relatively limited, and the model structures and their effectiveness still require further validation.

\subsection{SSMs for Remote Sensing Image Pan-Sharpening}
Pan-sharpening involves integrating low-resolution multi-spectral images (MSI) with high-resolution panchromatic (PAN) images to produce a high-resolution counterpart of the MSI. Pan-Mamba \cite{43} represents the first application of Mamba technology in remote sensing pan-sharpening, introducing channel exchange scanning and cross-modal scanning to enhance the efficient exchange and fusion of cross-modal information in image sharpening. Specifically, the first half of the features from both types of images are concatenated with the second half from the other type, and an SSMs gating mechanism is incorporated to facilitate feature fusion while suppressing redundant features. This feature fusion approach is also reflected in subsequent works, such as FMG \cite{124}, FMSR \cite{86}, and IRSRMamba \cite{87}.

The above methods in pan-sharpening achieve efficient cross-modal fusion through clever feature concatenation and the SSM gating mechanism. However, these methods may be highly dependent on the specific relationship between panchromatic and multispectral images, and adjustments may be needed when generalizing to other modal combinations.

\section{Advancements of SSMs Architecture Design in Remote Sensing}

This section will review advancements in SSMs architecture design for remote sensing. We will organize these advanced methods along two lines: 1) the model structure from the outside in, which includes high-level framework configurations, hybrid architecture designment, and the fundamental component refinement of SSM; 2) the progressive processing of data, which includes scanning strategies for feature extraction, feature fusion, as well as how to remove redundant features and how to handle frequency domain features.

\subsection{Scanning Strategies for Multi-Modal Remote Sensing Data}
\begin{figure*}
	\centering
	\includegraphics[width=1\linewidth]{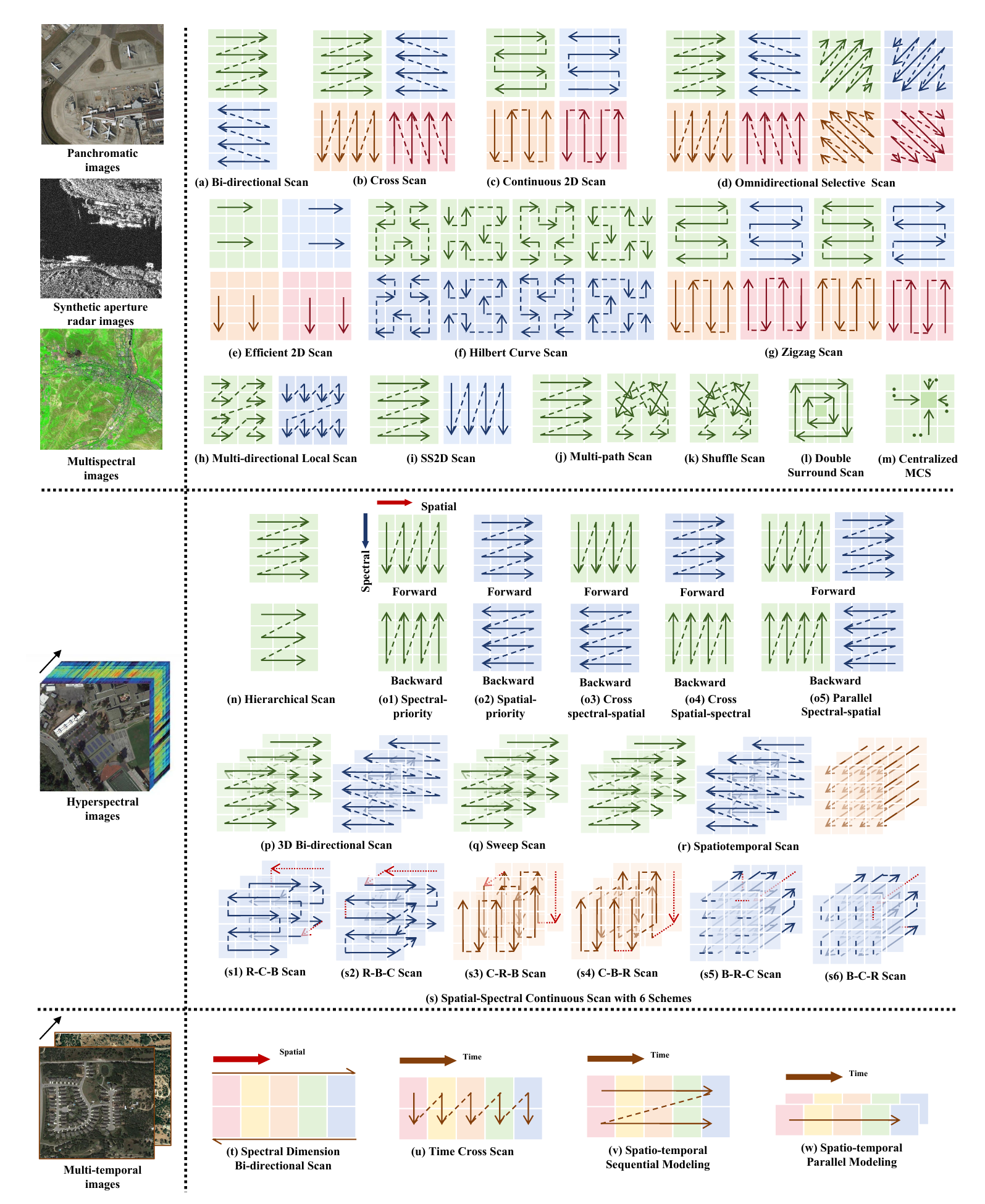}
        \vspace{-4ex}
	\caption{{Comparison of different SSM's scanning strategies in remote sensing. Methods (a) to (m) are applied to the processing of 2D remote sensing data, while methods (n) to (w) are applied to the processing of others, including hyperspectral imagery, time series data, and other forms of information.}}
	\label{fig:scanning}
\end{figure*}
The scanning strategies involve designing different scanning paths for the SSM. The purpose of the scanning strategy is to study specific model observation paths to achieve more efficient extraction of spatial and spectral information. The scanning mode adjusts the order of the input sequence and restores it to its original order after processing \cite{125,126,127}. 
Although some methods introduce multipaths, the increase in complexity still remains insignificant compared to the Transformer \cite{28,128}. Furthermore, when processing remote sensing data such as hyperspectral imagery, the scanning strategies can integrate spatial and spectral information effectively.

The reason of designing the scanning strategy for SSM is that remote sensing images lack inherent directionality. Relying on a single path sequence may introduce directional bias and lead to inadequate information integration in areas with non-uniform textures. In contrast, employing multi-path or adaptive sequences transforms the two-dimensional structure into a more balanced one-dimensional sequence, thereby reducing causal bias. This improvement enhances global consistency and the separability of small targets in complex remote sensing scenarios, as well as facilitates the efficient processing of high-dimensional data such as hyperspectral images (HSI). However, excessive scanning paths can easily introduce redundant information. When faced with diverse tasks and data modalities, choosing or combining the optimal scanning strategy poses a significant challenge.

We provide a detailed analysis of the advantages and disadvantages of the various methods, as seen in the Table \ref{tab:Comparisons_other_methods_scanning}. These advantages and disadvantages are derived from our analysis and summary. Additionally, Fig. \ref{fig:scanning} visually represents all 23 categories of scanning mechanisms employed in Mamba-based architectures for remote sensing tasks. Among these methods, (a)-(g) are regarded as conventional practices commonly employed in computer vision \cite{129,130,131}, with a primary emphasis on the continuity and directionality of scanning. In contrast, the methods from (h)-(n) place greater emphasis on scanning arrangement strategies, such as forward and reverse asymmetry (h) and (i), scale variation (n), randomness (j) and (k), and orientation (i)-(m). For multi-modal remote sensing data, the six methods labeled (o) explore asymmetric spectral and spatial forward and reverse sweeps. Methods (p)-(r) address specific spatial or spectral dimensions, while the six methods categorized as (s) alternately mix spatial and spectral information from the perspectives of arrangement and combination. Methods (t)-(w) focus on the evolving relationship between space and time.

\begin{table*}[!t]
\renewcommand{\arraystretch}{1.3}
\caption{{Comparison of advantages and disadvantages of typical scanning strategies in remote sensing}}
\label{tab:Comparisons_other_methods_scanning}
\centering

\resizebox{\textwidth}{!}{
\begin{tabular}{m{3cm}|m{7cm}|m{7cm}}

	\toprule

 \multicolumn{1}{m{3cm}|}{\shortstack{Scanning Method}} & {Strengths }& {Weaknesses}\\		
        \midrule
 \multicolumn{1}{c|}{(a) Bidirectional Scan\cite{154}} & {Compared to the simple forward and reverse directions, redundancy in information is reduced because of the asymmetry of the paths.} & {Artifacts or changes in resolution may lead to increased computational complexity.} \\
 \multicolumn{1}{c|}{(b) Cross Scan\cite{54} } & {Enhancing the capability to capture multi-directional spatial information.} & {The increase of computational complexity may necessitate additional steps to integrate information from cross-scans.}\\
 \multicolumn{1}{c|}{(c) Continuous 2D Scan\cite{129}}& {Enhancing spatial continuity, ensuring no jumps in the field of view.} & {The increase of computational complexity may not effectively capture the discrete spatial structure when the image is large.}\\
 \multicolumn{1}{c|}{(d) Omnidirectional Selective Scan\cite{45}}& {Aligned with the distinct physical characteristics of remote sensing images, regardless of the horizontal direction.} & {The increase of computational complexity may introduce redundant information.} \\
 \multicolumn{1}{c|}{(e) Efficient 2D Scan\cite{155}}& {Both local and global contexts are effectively captured while reducing computational costs.} & {Some patches are omitted, resulting in a loss of spatial continuity.}\\
 \multicolumn{1}{c|}{(g) Zigzag Scan\cite{156}}& {Providing comprehensive 2D spatial coverage while capturing both horizontal and vertical dependencies.} & {Increasing computational complexity}\\
 \multicolumn{1}{c|}{(h) Multi-directional Local Scan\cite{157}}& {Capturing fine-grained details and ensuring efficient handling of local regions.} & {Long-range dependencies may be overlooked, necessitating careful resizing of the window.} \\
 \multicolumn{1}{c|}{(i) SS2D Scan\cite{143}}& {Maintaining two-dimensional spatial dependencies by scanning in multiple directions to improve feature discriminability.} & {Increasing computational complexity.} \\
 \multicolumn{1}{c|}{(j) Multi-path Scan\cite{102}}& {The information from these different paths mitigates the effects of causal sequences.} & {Increasing computational complexity} \\
 \multicolumn{1}{c|}{(k) Shuffle Scan\cite{158}}& {Reduces the inductive bias caused by the causal scanning method during multi-modal fusion.} & {Weak to learn spatial continuity.} \\
 \multicolumn{1}{c|}{(l) Double Surround Scan\cite{159}}&{Prioritizing the modeling of relevant hierarchical environmental components enables the model to capture the spatial relationships of BEV features, enhancing its context-aware capability to better distinguish between foreground and background.} & {Weak to capture dependence on long-distant pixels} \\
 \multicolumn{1}{c|}{(n) Hierarchical Scan\cite{160}}& {Local and global features are captured at different scales, providing effective features for complex images.} & {Increasing model complexity and computational requirements necessitate careful design to balance local and global information.} \\
 \multicolumn{1}{c|}{(o) Spatial-Spectral Scan\cite{103}}& {Instead of employing a multi-directional scan, a new direction is used to replace the original reverse scan, reducing computational complexity} & {Weak to capture spatial and spectral information.} \\
 \multicolumn{1}{c|}{(p) 3D Bi-directional Scan\cite{154}}& {The spatial relationships between 3D data are captured to effectively process volumetric image data.} & {Capturing 3D information requires increased memory capacity.}\\
 \multicolumn{1}{c|}{(q) Sweep Scan\cite{161}}& {The bidirectional spatial dependencies of the HSI are effectively captured.} & {The long-sequence spectral dependencies are overlooked.} \\
 \multicolumn{1}{c|}{(s) Spatial-Spectral Continuous Scan\cite{138}}& {Preserving the spatial-spectral correlation within the local context and enhancing local texture exploration} & {The ability to capture long-distant spatial correlation is diminished.}\\
 \multicolumn{1}{c|}{(u) Time Cross Scan\cite{46}}& {Temporal cross-scanning enhances the understanding through time dimension.} & {Weak to capture spatial information.}\\

	\bottomrule
\end{tabular}}
\end{table*}

\subsection{High-level Framework Configurations}
High-level frameworks refer to strategies that incorporate SSM modules at a structural level. The main focus is to showcase the model's construction approach. In the field of remote sensing, these variations can be classified into two categories: Backbone-Centric methods and U-Net-based methods.

Backbone-Centric methods focus on feature extraction pipelines \cite{27,47,56,57,58,59,71,80,90,102,132,133,134,135,136,137,138,139,140,141,142}, these approaches embed SSMs as core computational units in backbone networks. Key innovations include: 1) Geometric-aware token scanning strategies for remote sensing data alignment\cite{56,59,90,102}; 2) Dynamic feature weighting mechanisms for noise suppression\cite{57,58,138}; 3) Hybrid designs integrate
multi-scale modules (e.g., pyramid frameworks) to compensate
for SSM’s local feature limitations \cite{45,76,78,85}. 

U-Net-based methods integrate SSMs modules into U-Net architectures through two primary strategies \cite{46,77,81,86,87,91,92,123,143,144,145}: 1) The first strategy retains the complete U-Net\cite{114,146} framework but replaces its central computational layers with SSM blocks, as seen in designs using residual blocks with integrated SSMs \cite{93}; 2) The second strategy deploys U-Net components for specialized purposes, such as parallel U-Net branches for spatial/spectral feature isolation \cite{123}, and combining SSMs with convolutional layers in skip connections to refine multi-scale fusion \cite{147}.

Overall, Backbone-Centric methods treat SSM as an efficient feature extractor, with a design that prioritizes the construction of hierarchical feature representations. Although structures such as FPN can be added to obtain multi-scale features, their feature fusion mechanisms are typically not as native and straightforward as those of U-Net-like methods. U-Net-based methods can better integrate multi-scale information, but designing complex structures, such as parallel branches, can add additional complexity and training difficulties for SSM.

\begin{figure*}
	\centering
	\includegraphics[width=1\linewidth]{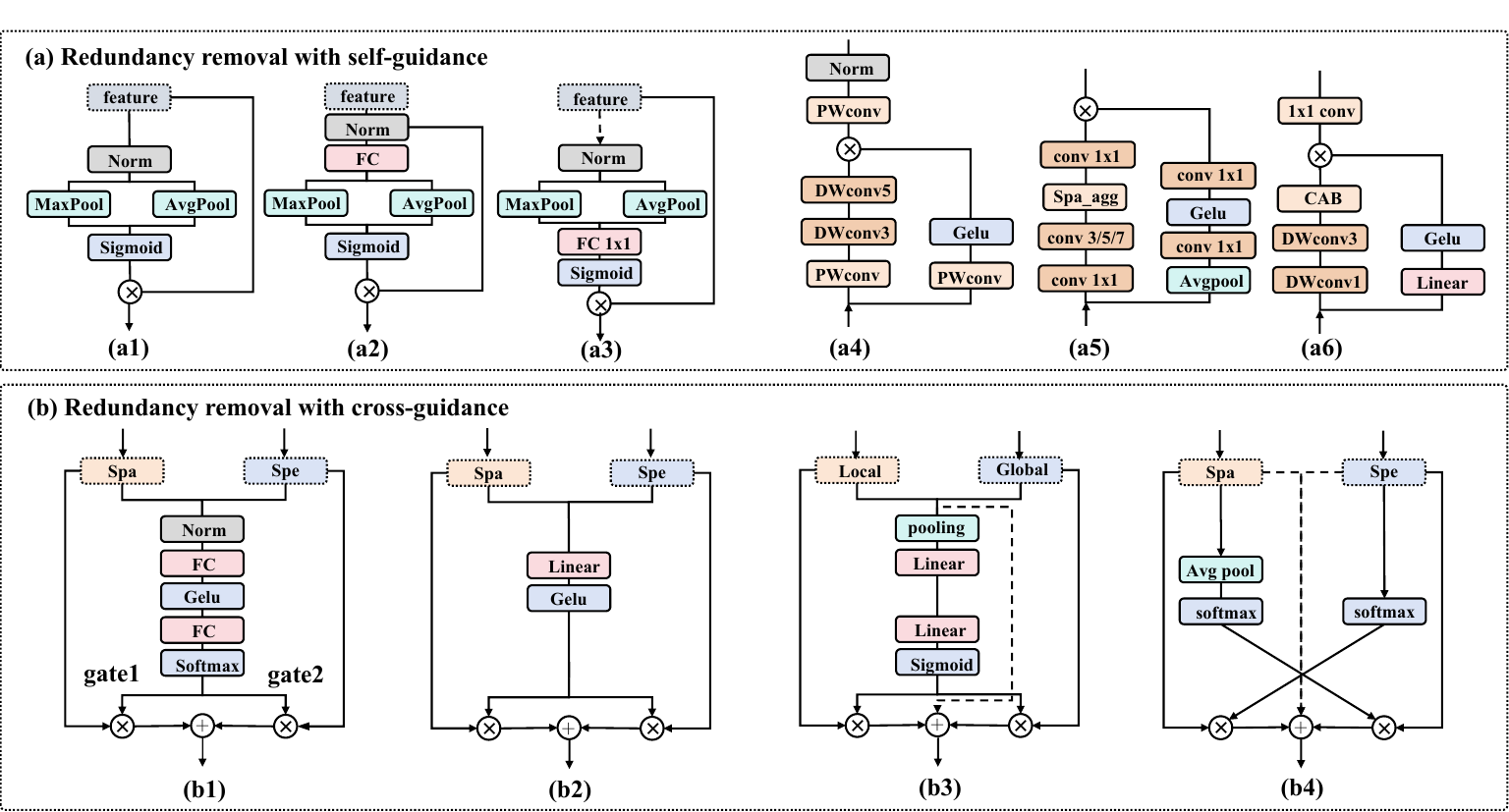}
        \vspace{-4ex}
	\caption{{Comparison of the redundancy removal methods, including self-guidance and cross-guidance, where spa and spe represent spatial and spectral features, respectively.
}}
	\label{fig:feature}
\end{figure*}

\subsection{Hybrid Architecture Designment}

The hybrid architecture designment aims to combine SSMs with classical structures such as CNN and Transformer, either in parallel, serial, or nested ways, in order to compensate shortcoming of pure SSMs. In this section, we review hybrid architectures that combine SSMs with other powerful techniques: CNN for enhanced local understanding, Transformer for enhanced global understanding, and GNN for graph-structured data.

In SSM-CNN hybrids, convolutional inductive biases enhance local feature extraction, which serves as a critical supplement to SSMs' weaker spatial awareness\cite{44,77,86,92,93,138,148,149,150,151,152,153}. Techniques like depthwise convolution (DW) for multi-scale perception \cite{56,69} and pointwise convolution (PW) for cross-modal interaction \cite{162} are systematically employed. This can be seen in HSIMamba’s \cite{57} hierarchical feature refinement and SpectralMamba’s \cite{162} dynamic gated spectral-spatial fusion. Meanwhile, channel attention mechanisms and dual-branch designs have become ubiquitous in advanced frameworks. This is exemplified by DualMamba’s CNN-SSM co-training \cite{63} and SSFNet’s DW-PW cross-modal fusion \cite{162}. Another method, such as Pyramid-Mamba\cite{163}, uses SSM to process long-distance or multi-scale features, enabling effective extraction and fusion of global contextual information. After that, CNN is applied to further process these features, capturing fine-grained spatial characteristics.

In SSM-Transformer hybrids, a common approach is to explore mixed architectures, which combine the global modeling capabilities of the self-attention mechanism with the selective information acquisition of SSM: MamTrans adds a Transformer branch outside the order-sensitive SSM branch, both extract features in parallel\cite{62}; while MHSSMamba\cite{66} and RS3Mamba\cite{76} synergize spectral-spatial attention with SSMs’ long-sequence dependencies for enhanced segmentation granularity. Architectures like MambaFormerSR further innovate through nested SSM-attention structures and sparse cross-attention variants \cite{81,88,89,90,164,165,166}. 

In SSM-GNN hybrids, Yang \textit{‌et al.} proposed GraphMamba, which replaces the residual normalization layer and multi-head attention mechanism layer of ViT with SSM and graph network respectively \cite{64}. A filter matrix that changes with the nodes of the graph is designed to better represent the global information and abstract the relationship between different nodes, which is conducive to the extraction of spatial \cite{65,167,168,169,170}.

The purpose of the hybrid architecture of SSM is to balance the long-range context of remote sensing images with the texture and boundaries of fine-grained objects. In this framework, SSM is responsible for long-range modeling, while CNNs, deformable convolutions, or sparse attention mechanisms focus on local refinement. This approach enables the model to simultaneously capture both global and local information effectively. This cross-paradigm integration systematically addresses SSMs’ limitations while preserving their efficiency advantages, establishing hybrid architectures as the dominant paradigm in contemporary remote sensing SSMs research. Furthermore, effectively integrating models of different paradigms while balancing their contributions poses a significant design challenge. Additionally, the introduction of complex modules like Transformers may partially negate the original efficiency advantages of SSM, requiring careful design to maintain high efficiency. Moreover, the internal working mechanisms of hybrid architectures and the interactions between different components still need more in-depth theoretical research.

\subsection{Fundamental Component Refinement}

\begin{figure*}
	\centering
	\includegraphics[width=1\linewidth]{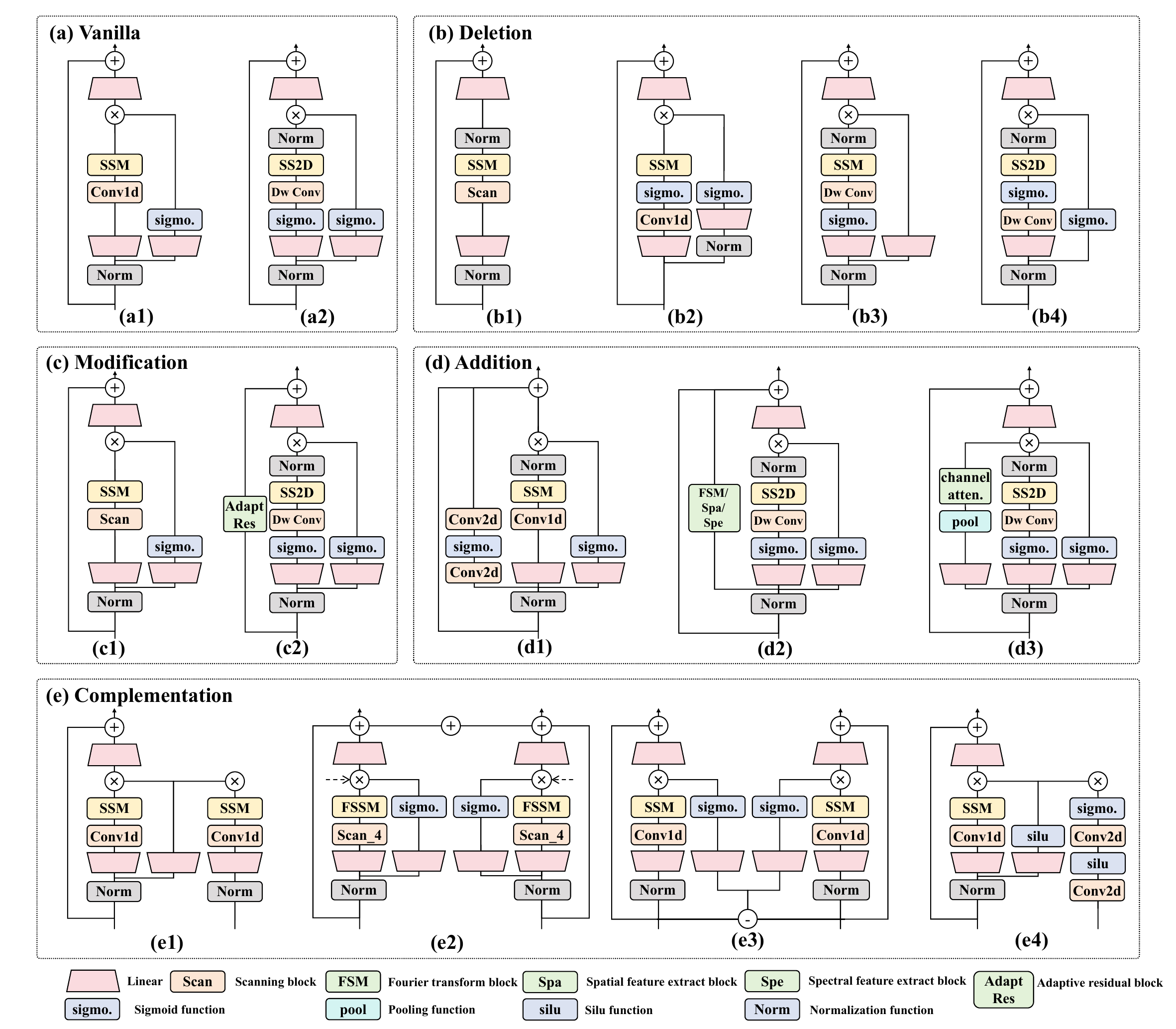}
        \vspace{-4ex}
	\caption{{Comparison of various methods for internal deformation of SSM. In panel (a), the vanilla Mamba and its second generation are presented, while panels (b), (c), (d), and (e) represent the four specific methods of adjusting the structure of SSM: deletion, modification, addition, and complementation.}
 }
	\label{fig:internal-deformation}
\end{figure*}
Fundamental component refinement represents an enhancement that begins with the smallest structural unit of the network. It involves adjusting the structure of SSMs through four methods: deletion, modification, addition, and complementation. 

Deletion aims to reduce the computational load by removing unnecessary components, such as linear projection layers. As shown in Fig. \ref{fig:internal-deformation}, models such as ss-mamba (Fig. \ref{fig:internal-deformation} b2) \cite{59}, RS3mamba (Fig. \ref{fig:internal-deformation} b3) \cite{76}, and RSdehamba (Fig. \ref{fig:internal-deformation} b4) \cite{92} have implemented similar strategies. Additionally, since mamba architectures are often integrated with other structures, to streamline the architecture, internal normalization layers can be removed if external normalization is already applied.

Modification focuses on altering internal layers. For instance, GraphMamba (Fig. \ref{fig:internal-deformation} c2) \cite{65} replaces the skip connection with an adaptive residual branch, which enhances the model's capability to extract features. Besides, changing the scan order is categorized as a variant of the scan strategy mentioned earlier (Fig. \ref{fig:internal-deformation} c1).

Addition involves incorporating an extra branch into the existing framework, introducing a new trainable computational module to better manage the original information. \cite{72,73,171,172}. For example, Cdmamba (Fig. \ref{fig:internal-deformation} d1) \cite{27} introduces an additional two-dimensional convolutional and activation branch, while FMSR (Fig. \ref{fig:internal-deformation} d2) \cite{86} enhances the learning of spectral features by integrating Fourier transform branches. Additionally, GMSR (Fig. \ref{fig:internal-deformation} d3) incorporates both spectral and spatial branches \cite{145}.

Complementation involves providing supplementary inputs \cite{158}. When the newly added modal information is removed, we can observe that this type of SSM can be approximated as a vanilla SSM. For instance, Pan-mamba (Fig. \ref{fig:internal-deformation} e1) \cite{43} and Fusionmamba (Fig. \ref{fig:internal-deformation} e2) \cite{123} leverage the differences among them or the characteristics of various modalities and visual field information to enhance and complement each other's data.

Fundamental component refinement is quite challenging because improper designment may disrupt the flow of critical information in the model, impacting performance. Guiding the refinement of SSMs from a theory-driven perspective could offer a more reliable and promising path for future development.

\subsection{Feature Fusion of Multi-Modal Remote Sensing Data}

Multi-modal fusion encompasses pixel-level, feature-level, and decision-level integration. In the context of SSM, the current emphasis is on feature-level fusion. This is because SSMs places greater importance on global spatial information, and such feature fusion can supplement SSMs with more spectral/multi-modal information.  Pixel-level and decision-level fusion require further research. The feature-level fusion approaches can be categorized into complementary use of data from different time periods, spatial features, and spectral features, as well as the integration of information from different modalities \cite{43,69,85,158}. Table \ref{tab:Comparisons_other_methods_fusion} summarizes the information categories, and main idea used by various models\cite{173,174,175,176,177,178,179}.

The purpose of multi-modal fusion in SSM is to address the challenges of cross-modal alignment and redundant information filtering in remote sensing images, where noise is easily amplified post-fusion\cite{180,181,182}. By employing cross-modal guided selective gating and dedicated scanning paths, the approach suppresses channels or regions with low mutual information while enhancing complementary frequency bands. However, this method relies on alignment quality and hyperparameter tuning, resulting in a training cost that is higher than that of single-modal approaches. In addition, under the survey framework adopted in this paper, the multimodal fusion mentioned in Section 3 specifically refers to methods that arise in particular tasks, without focusing on data modalities, but rather on task categories. In contrast, the multimodal fusion discussed in this section inclined more towards focusing on fusion methods rather than the tasks themselves.

\subsection{Gated Redundancy Removal}

In multi-directional scanning of remote sensing data, the combination of multi-modal data and various hybrid structures inevitably results in the generation of redundant information \cite{183,184}. As a result, many studies in SSMs employ a gating mechanism to filter out redundant information, a process we named gated redundancy removal. 

Gated redundancy removal can be either conducted within the main branch or the auxiliary branches of the networks. The extracted features are simultaneously fed into two branches, where the auxiliary branch utilizes dual pooling, filtering, or mixing with other modalities for further processing. Meanwhile, the main branch decides which parts of the extracted features to retain or discard, based on signals from the auxiliary branches. At the junction where the main path and the auxiliary branches converge, a gated element-wise multiplication is typically implemented. In addition, a threshold can be introduced to further enhance the information screening process, filtering out unnecessary information. Based on the characteristics of the auxiliary branches, we categorize these into two specific approaches, including self-guidance and cross-guidance, as shown in Fig. \ref{fig:feature}. The purpose of gated redundancy removal is to reduce significant redundancy and ineffective calculations brought about by large-scale or multi-modal inputs. This can decrease computational load and memory usage, maintain performance in key areas, and enhance the efficiency of remote sensing interpretation. However, the selection of features and the setting of thresholds have a significant impact on the performance, necessitating careful tuning.

Self-guidance focuses on redundancy removal within a single stream by comparing different transforms of the same feature. Such methods commonly utilize a variety of dual feature processing techniques, including average pooling versus max pooling, as well as CNN features from different scales. For example, GMSR's \cite{138} spatial gradient attention (Fig. \ref{fig:feature} a1), spectral gradient attention (Fig. \ref{fig:feature} a3), and M-CD (Fig. \ref{fig:feature} a2) \cite{122} utilize average pooling and max-pooling operations along the channel dimension to effectively address channel information. Le-Mamba's MSCGU modules (Fig. \ref{fig:feature} a4) \cite{105}, CM-UNet (Fig. \ref{fig:feature} a5) \cite{77}, and FMSR (Fig. \ref{fig:feature} a6) \cite{86} utilize CNNs with varying kernel sizes within the main branch to aggregate information from different receptive fields, enabling the capture of a wide range of features.

\begin{table}[!t]
\renewcommand{\arraystretch}{1.3}
\caption{{Comparison of different methods for feature fusion of multi-modal remote sensing data}}
\label{tab:Comparisons_other_methods_fusion}
\centering

\resizebox{0.9\linewidth}{!}{
\begin{tabular} {c|c|c}
	\toprule

\multicolumn{1}{c|}{Method} & {Category} & {Main idea}\\
        \midrule

\multicolumn{1}{c|}{	SSRFN\cite{185}	} & {	HSI/MSI	} & {Constructing a mamba cross-attention network to fuse features} \\
\multicolumn{1}{c|}{	Msfmamba\cite{69}	} & {	HSI/LiDAR	} & {Designing a cross-guided mamba for fusion} \\
\multicolumn{1}{c|}{	Shufflemamba\cite{158}	} & {	LRMS/PAN	} & {Designing a Cross-guided mamba for fusion} \\
\multicolumn{1}{c|}{	Pan-mamba\cite{43}	} & {	IRMS/PAN	} & {Designing a Cross-guided mamba for fusion} \\
\multicolumn{1}{c|}{	DMM\cite{44}	} & {	MSI/RGB	} & {Designing a Cross-guided mamba for fusion} \\
\multicolumn{1}{c|}{	TFFNet\cite{179}	} & {HSI/SAR} & {Designing a squeeze-and-excitation transform fusion module} \\
\multicolumn{1}{c|}{	RSM\cite{45}	} & {bi-temporal} & {Performing element-wise addition on the features and decode} \\
\multicolumn{1}{c|}{	Changemamba\cite{46}	} & {bi-temporal} & {Concatenating the multi-temporal features and decode using SSM} \\
\multicolumn{1}{c|}{	RSCama\cite{85}	} & {bi-temporal} & {Sorting tokens in a specific order and then processing through SSM} \\
\multicolumn{1}{c|}{	M-CD\cite{122}	} & {bi-temporal} & {Concatenating the multi-temporal features and decode using SSM} \\
\multicolumn{1}{c|}{	Le-mamba\cite{105}	} & {	spectral/spatial	} & {Fusing the features through element-wise multiplication} \\
\multicolumn{1}{c|}{	Fusionmamba\cite{123}	} & {	spectral/spatial	} & {Designing a Cross-guided mamba for fusion} \\
\multicolumn{1}{c|}{	Ss-mamba\cite{59}	} & {	spectral/spatial	} & {Modulating spatial and spectral tokens using center region information} \\
\multicolumn{1}{c|}{	GMSR\cite{138}	} & {	spectral/spatial	} & {Concatenating the spatial and spectral features} \\
\multicolumn{1}{c|}{	MamTrans\cite{62}	} & {	spectral/spatial	} & {Concatenating the spatial and spectral features} \\
\multicolumn{1}{c|}{	Dualmamba\cite{63}	} & {	spectral/spatial	} & {Designing a Cross-guided mamba for fusion} \\
\multicolumn{1}{c|}{	MHSSMamba\cite{66}	} & {	spectral/spatial	} & {Constructing a cross-attention network to fuse features} \\
\multicolumn{1}{c|}{	Morpmamba\cite{106}	} & {	spectral/spatial	} & {Constructing a cross-attention network to fuse features} \\

	\bottomrule
\end{tabular}
}
\end{table}

Cross-guidance utilizes information from another stream to guide redundancy removal, such as from a different modality or scale. Leveraging the differences between these modalities and scales facilitates a comprehensive control feature process. Specifically, S$^2$Mamba (Fig. \ref{fig:feature} b1) \cite{58} filters out low-contribution features by artificially defining a threshold at the gate. This gating mechanism effectively truncates redundant features that may hinder HSI classification while promoting the integration of spatial and spectral attributes. SS-Mamba (Fig. \ref{fig:feature} b2)  \cite{59} utilizes the average value of spectral information and the intermediate value of spatial information to filter out the low-contribution features of the opponent. Dualmamba (Fig. \ref{fig:feature} b3) \cite{63} and Spectral-Spatial Mamba (Fig. \ref{fig:feature} b4) \cite{57} employ pooling methods to achieve a broader range of feature capture.

\subsection{Processing in Frequency Domain}

Remote sensing images inherently contain noise and distortions introduced by complex acquisition conditions, such as atmospheric scattering and sensor noise. These interferences manifest in the frequency domain as anomalous energy distribution patterns, such as random noise in the high-frequency range or smooth distortions in the low-frequency range\cite{186}. SSMs can leverage its long-sequence modeling capabilities to learn the relationships between multiple frequency ranges. Therefore, SSMs frameworks increasingly integrate domain-specific transformations to address these challenges\cite{86,87,89}. FMSR \cite{86} embeds Fourier transforms in SSM block to decompose global-local spectral features through frequency-domain analysis, while MambaFormerSR enhances this paradigm with frequency band attention for adaptive feature refinement \cite{89}. Extending beyond spectral processing, IRSRMamba employs wavelet transforms to capture multi-directional texture patterns, effectively generalizing frequency-based approaches to spatial-textural domains \cite{87}. WMSR\cite{187} uses discrete wavelet transform to separate features into low-frequency and high-frequency components in the frequency domain, and processes the low-frequency feature with SSM. These methods enable SSMs to simultaneously focus on spatial and frequency domain features.

\section{Challenges, Opportunities and Limitations}

In this section, we will discuss the challenges, opportunities, and limitations, including the SSM-based remote sensing foundation model, efficient SSMs for edge devices and large-scale image processing, SSM-driven remote sensing vision-language model, and limitations of applying SSMs in remote sensing.

\subsection{SSM-based Remote Sensing Foundation Model}

The foundation model represents a new trend in the development of remote sensing, defined as models with billions of parameters, pre-trained on large-scale datasets, and capable of adapting to various downstream tasks. However, current methods employ Transformer architectures rather than SSMs, and they face significant limitations. The performance of the Transformer diminishes in remote sensing applications with numerous spatially dispersed small targets\cite{188}. While increasing the input resolution can mitigate this issue, it leads to high costs due to quadratic growth in GPU consumption and computational complexity\cite{188,189,190,191,192,193,194,195,196,197,198,199,200,201,202,203}. This drastically reduces the efficiency of remote sensing foundation models in tasks such as wide area data processing and dense prediction. In contrast, the structure of SSMs is designed for long-distance modeling and features approximately linear computational complexity, making it inherently suitable for these tasks. At the same time, SSMs possess selective modeling capabilities, making them inherently well-suited for the detection of sparse targets in remote sensing. Therefore, designing an SSM-based remote sensing foundation model is crucial\cite{12,204,205,206}.

The challenges and opportunities of SSM-based remote sensing foundation models can be summarized in three key points. Firstly, scaling to global-scale datasets is essential for remote sensing. This can help remote sensing models generalize across the globe rather than overfitting to a few regions. It ensures that RS-SSMs can support mission-critical applications such as deforestation monitoring, maritime domain awareness, and disaster response on a planetary scale. Secondly, current models often suffer from the issue of long sequence forgetting, particularly in scenarios involving sparse small targets or densely arranged repetitive targets in remote sensing images. Thus, exploring mechanisms for context scanning or fusion methods designed specifically for ultra-long distances is crucial for advancing foundational models. One feasible approach is to leverage SSM's selective information capture and selective activation of parameters to construct a remote sensing backbone model with long-range modeling capabilities, making it particularly suited for tasks such as ship detection on global-scale datasets. Thirdly, as the number of parameters in foundation models and the amount of data increase rapidly, pretraining with fine-tuning is considered a viable approach. However, existing mainstream efficient fine-tuning methods, such as adapters and LoRA, do not support training for SSM models, indicating a significant need for further research\cite{207,208,209,210,211}. Current work has demonstrated that SSM has the capability for large-scale data learning\cite{188}. It is important to explore efficient pre-training and fine-tuning techniques specifically for SSM. For example, assigning an increment to the parameter matrix of SSM during fine-tuning and allowing only the increment to be learned could be beneficial.

\subsection{Efficient SSMs for Edge Devices and Large-scale Image Processing}

In many remote sensing scenarios, particularly those involving large-scale image processing and edge devices, efficient models have always garnered significant attention. The SSMs possess natural advantages. The existing challenges and opportunities can be summarized as follows.

In SSMs' acceleration techniques, there have been several studies on acceleration methods for 2D image processing in the field of computer vision\cite{199}. However, remote sensing data is more complex and diverse. SSMs were designed with hardware acceleration at the beginning\cite{1}, and there has already been progress in further accelerating SSMs for hardware\cite{212}. The research on acceleration techniques for multi-modal remote sensing data (such as HSI, MSI, text data, and SAR) holds great promise, especially when it fully leverages hardware efficiency \cite{213,214}. One potential area of exploration is the development of transmission algorithms that enable the adaptation of 2D and 3D data to the I/O layer of SSM. In this context, transmission algorithms refer to the process where data is transferred between the CPU and GPU during certain matrix operations. By designing specific operators, the computation order after uploading to the GPU can be altered to reduce the transfer time between different hardware.

At the same time, model lightweight techniques, including distillation, pruning, and quantization, have found various applications in remote sensing scenarios\cite{215,216,217}. Since methods such as pruning and distillation need to be designed based on the structural characteristics of different models, methods tailored specifically for the structure of SSMs and particular types of remote sensing data remain to be further explored \cite{218,219,220,221,222,223,224}. In the future, in lightweight methods such as distillation, introducing simulated remote sensing noise as interference could compel the SSM, functioning as the student model, to concurrently learn its adaptability to remote sensing images. Additionally, pruning and quantization can be utilized to enhance the inference speed of the SSM, making it suitable for time-sensitive practical deployment scenarios in remote sensing. Moreover, by designing new operators that enhance the parallelism of the internal state equations and output equations of SSM in conjunction with specific hardware, it could be better suited for deployment in resource-constrained remote sensing scenarios.

\subsection{SSM-driven Remote Sensing Vision-Language Model}

Currently, in the field of visual-language models specifically for remote sensing, contrastive models and autoregressive models are the two dominant approaches. SSMs present both advantages and challenges in relation to these two models.

Contrastive models like CLIP align the embeddings of images and text through contrastive learning. In the context of remote sensing, the complexity of the scenes and their broad coverage require lengthy textual descriptions. However, not every word in the long text description is relevant to the corresponding image (such as adjectives or background information).  CLIP's text encoder, which is based on Transformers, treats all tokens equally and lacks the ability to dynamically filter them. In contrast, SSMs possesses the capability for selective modeling of information, and its linear complexity makes it well-suited for handling long text descriptions in remote sensing applications. Research on SSMs in this area is relatively sparse, and how to achieve effective modality alignment warrant further investigation.

Autoregressive models are currently the most common architecture in multi-modal generative modeling \cite{225,226,227,228}. These models primarily rely on Transformer backbones, leading to a high computational complexity. This issue is particularly pressing in the current era of large models, where today's Vision-Language Models (VLMs) possess a significant number of parameters and computational demands. Furthermore, the complexity of remote sensing scenarios, which necessitates the generation of long detailed texts, further exacerbates the computational burden. In contrast, SSMs, as a type of efficient sequence modeling framework, are inherently well-suited for autoregressive architectures for generating long detailed texts. However, there is limited existing research on the design of SSM architectures for autoregressive modeling, specifically in the context of remote sensing. Recently, Deepseek-OCR\cite{229} has explored the feasibility of processing text as images, achieving a tenfold model compression. Similarly, SSM also possesses efficient information summarization capabilities. The exploration of SSM as a purpose encoder for large-scale remote sensing images and textual data represents a viable approach.

\subsection{Limitations of Applying SSMs in Remote Sensing}
Despite the rapid development of RS-SSMs, there are still limitations and open issues in applying SSMs in remote sensing, including scale variation, fine-grained texture modeling, and long-sequence stability.

SSMs are challenged first by the extreme scale variation and strong geometric diversity in large remote sensing images, which makes purely sequential modeling difficult to consistently excel. Many SSM-based designs still rely on additional locality and multi-scale priors to achieve strong dense prediction performance\cite{230}. However, it remains unclear when a pure SSM backbone is preferable to a hybrid design, and how to combine SSMs with local operators or multi-scale mechanisms without sacrificing efficiency, remains an open problem.

SSMs also face open questions regarding long-sequence stability and memory behaviors, especially when remote sensing images are converted into extremely long sequences. For very large images, sparse selection mechanisms in SSMs may introduce information loss or instability as sequence length grows, and robustness across different sensors, resolutions, and acquisition conditions remains insufficiently understood. Addressing these limitations requires future SSM research to more systematically evaluate stability under ultra-high-resolution remote sensing images.

SSMs further encounter limitations in capturing fine-grained textures and high-frequency details that are crucial for remote sensing interpretation, such as building edges, road boundaries, and subtle material patterns in HSI. Scanning strategies and long-range mixing in SSMs alone may be less sensitive to these local texture cues, motivating the integration of CNN-like local operators, frequency-domain processing, or explicit multi-scale feature fusion within SSM frameworks. However, a systematic understanding is still missing regarding which augmentation best complements SSMs under different tasks, modalities, and spatial resolutions.

\section{Conclusion}

Since the introduction of SSMs in remote sensing in 2024, a series of advanced research studies have emerged. This paper extensively reviews the SSMs applications in remote sensing tasks, advancements in architecture design, existing challenges, and future opportunities. With the rapid development of technology today, we believe this paper can help researchers quickly grasp the development context of SSMs in remote sensing and the key ideas of model improvement, and better integrate into the future direction in this field.

	\Acknowledgements{The work was supported by the National Natural Science Foundation of China (Grant Nos. 62125102, 62471014, U24B20177 and 624B2017), and the Fundamental Research Funds for the Central Universities.}


\begin{thebibliography}{226}

\bibitem{1} Gu A, Dao T. Mamba: Linear-time sequence modeling with selective state spaces. In First conference on language modeling, 2024
\bibitem{2} Zhang H, Gong Y, Shen Y, et al. Poolingformer: Long document modeling with pooling attention. International Conference on Machine Learning, 2021: 12437–12446
\bibitem{3} Dao T, Fu D, Ermon S, et al. Flashattention: Fast and memory-efficient exact attention with io-awareness. Advances in Neural Information Processing Systems, 2022, 35: 16344–16359
\bibitem{4} Wu C, Wu F, Qi T, et al. Fastformer: Additive attention can be all you need. ArXiv preprint arXiv:2108.09084, 2021
\bibitem{5} Beltagy I, Peters M E, Cohan A. Longformer: The long-document transformer. ArXiv preprint arXiv:2004.05150, 2020
\bibitem{6} Ma W, Chen C, Ma M, et al. An adaptive dual-supervised cross-deep dependency network for pixel-wise classification. IEEE Transactions on Geoscience and Remote Sensing, 2025
\bibitem{7} Chen K, Liu C, Chen H, et al. Rsprompter: Learning to prompt for remote sensing instance segmentation based on visual foundation model. IEEE Transactions on Geoscience and Remote Sensing, 2024
\bibitem{8} Zhu W, Guo L, Zhang T, et al. Fastformer: transformer-based fast reasoning framework. Fourteenth International Conference on Graphics and Image Processing (ICGIP 2022), 2023, 12705, 378–387
\bibitem{9} Cheng B, Misra I, Schwing A G, et al. Masked-attention mask transformer for universal image segmentation. Proceedings of the IEEE/CVF conference on computer vision and pattern recognition, 2022: 1290–1299
\bibitem{10} Zou Z, Chen K, Shi Z, et al. Object detection in 20 years: A survey. Proceedings of the IEEE, 2023, 111: 257–276
\bibitem{11} Cheng G, Han J. A survey on object detection in optical remote sensing images. ISPRS journal of photogrammetry and remote sensing, 2016, 117: 11–28
\bibitem{12} Awais M, Naseer M, Khan S, et al. Foundation models defining a new era in vision: a survey and outlook. IEEE Transactions on Pattern Analysis and Machine Intelligence, 2025
\bibitem{13} Redmon J. You only look once: Unified, real-time object detection. Proceedings of the IEEE conference on computer vision and pattern recognition, 2016
\bibitem{14} Redmon J, Farhadi A. Yolo9000: better, faster, stronger. Proceedings of the IEEE conference on computer vision and pattern recognition, 2017: 7263–7271
\bibitem{15} Liu W, Anguelov D, Erhan D, et al. Ssd: Single shot multibox detector. Computer Vision–ECCV 2016: 14th European Conference, 2016: 21-37
\bibitem{16} Krizhevsky A, Sutskever I, Hinton G E. Imagenet classification with deep convolutional neural networks. Advances in neural information processing systems, 2012, 25
\bibitem{17} He K, Zhang X, Ren S, et al. Deep residual learning for image recognition. Proceedings of the IEEE conference on computer vision and pattern recognition, 2016: 770–778
\bibitem{18} Sun X, Weng X, Pang C, et al. Mitigating representation bias for class-incremental semantic segmentation of remote sensing images. Science China Information Sciences, 2025, 68: 182301
\bibitem{19} Gu A, Goel K, R{\'e} C. Efficiently modeling long sequences with structured state spaces. ArXiv preprint arXiv:2111.00396, 2021
\bibitem{20} Bansal S, Madisetty S, Rehman M Z U, et al. A comprehensive survey of mamba architectures for medical image analysis: Classification, segmentation, restoration and beyond. ArXiv preprint arXiv:2410.02362, 2024.
\bibitem{21} Qu S, Tao X, Qu Z, et al. Almrr: Anomaly localization mamba on industrial textured surface with feature reconstruction and refinement. Chinese Conference on Pattern Recognition and Computer Vision (PRCV), 2024: 378–391
\bibitem{22} Tian W, Zeng H, Zhao Y P, et al. Empowering snapshot compressive imaging: Spatial-spectral state space model with across-scanning and local enhancement. ArXiv preprint arXiv:2408.00629, 2024
\bibitem{23} Li H, Hu Q, Yao Y, et al. Cfmw: Cross-modality fusion mamba for multispectral object detection under adverse weather conditions. ArXiv preprint arXiv:2404.16302, 2024
\bibitem{24} D{\'\i}az A H, Davidson R, Eckersley S, et al. E-mamba: Using state-space-models for direct event processing in space situational awareness. IAA Conference on AI in and for Space (SPAICE), 2024: 509-514
\bibitem{25} Teng Y, Wu Y, Shi H, et al. Dim: Diffusion mamba for efficient high-resolution image synthesis. ArXiv preprint arXiv:2405.14224, 2024
\bibitem{26} Li K, Li X, Wang Y, et al. Videomamba: State space model for efficient video understanding. European Conference on Computer Vision, 2024: 237–255
\bibitem{27} Zhang H, Chen K, Liu C, et al. Cdmamba: Incorporating local clues into mamba for remote sensing image binary change detection. IEEE Transactions on Geoscience and Remote Sensing, 2025
\bibitem{28} Zhu Q, Fang Y, Cai Y, et al. Rethinking scanning strategies with vision mamba in semantic segmentation of remote sensing imagery: an experimental study. IEEE Journal of Selected Topics in Applied Earth Observations and Remote Sensing, 2024
\bibitem{29} Rahman M M, Tutul A A, Nath A, et al. Mamba in vision: A comprehensive survey of techniques and applications. ArXiv preprint arXiv:2410.03105, 2024
\bibitem{30} Sun X, Tian Y, Lu W, et al. From single-to multi-modal remote sensing imagery interpretation: A survey and taxonomy. Science China Information Sciences, 2023, 66: 140301
\bibitem{31} Bansal S, Madisetty S, Rehman M Z U, et al. A comprehensive survey of mamba architectures for medical image analysis: Classification, segmentation, restoration and beyond. ArXiv preprint arXiv:2410.02362, 2024
\bibitem{32} Lenz B, Lieber O, Arazi A, et al. Jamba: Hybrid transformer-mamba language models. The Thirteenth International Conference on Learning Representations, 2025
\bibitem{33} Zhao H, Zhang M, Zhao W, et al. Cobra: Extending mamba to multi-modal large language model for efficient inference. Proceedings of the AAAI Conference on Artificial Intelligence, 2025, 39: 10421–10429
\bibitem{34} Zeng F, Liu R, Sun X, et al. Multi-static isac based on network-assisted full-duplex cellfree networks: Performance analysis and duplex mode optimization. Science China Information Sciences, 2025, 68: 150303
\bibitem{35} Liu N, Li W, Wang Y, et al. A survey on hyperspectral image restoration: From the view of low-rank tensor approximation. Science China Information Sciences, 2023, 66: 140302
\bibitem{36} Gu Y, Liu T, Gao G, et al. Multimodal hyperspectral remote sensing: An overview and perspective. Science China Information Sciences, 2021, 64: 121301
\bibitem{37} Ma X, Zhang X, Pun MO, et al. A Unified Framework with Multimodal Fine-tuning for Remote Sensing Semantic Segmentation. IEEE Transactions on Geoscience and Remote Sensing, 2025
\bibitem{38} Qu H, Ning L, An R, et al. A survey of mamba. ArXiv preprint arXiv:2408.01129, 2024
\bibitem{39} Xu R, Yang S, Wang Y, et al. A survey on vision mamba: Models, applications and challenges. ArXiv preprint arXiv:2404.18861, 2024
\bibitem{40} Zhang H, Zhu Y, Wang D, et al. A survey on visual mamba. Applied Sciences, 2024, 14: 5683
\bibitem{41} Dosovitskiy A. An image is worth 16x16 words: Transformers for image recognition at scale. ArXiv preprint arXiv:2010.11929, 2020
\bibitem{42} Zhu L, Liao B, Zhang Q, et al. Vision mamba: Efficient visual representation learning with bidirectional state space model. ArXiv preprint arXiv:2401.09417, 2024
\bibitem{43} He X, Cao K, Zhang J, et al. Pan-mamba: Effective pan-sharpening with state space model. Information Fusion, 2025, 115: 102779
\bibitem{44} Zhou M, Li T, Qiao C, et al. Dmm: Disparity-guided multispectral mamba for oriented object detection in remote sensing. IEEE Transactions on Geoscience and Remote Sensing, 2025
\bibitem{45} Zhao S, Chen H, Zhang X, et al. Rs-mamba for large remote sensing image dense prediction. IEEE Transactions on Geoscience and Remote Sensing, 2024
\bibitem{46} Chen H, Song J, Han C, et al. Changemamba: Remote sensing change detection with spatio-temporal state space model. IEEE Transactions on Geoscience and Remote Sensing, 2024
\bibitem{47} Paranjape J N, Melo C D, Patel V M. A mamba-based siamese network for remote sensing change detection. 2025 IEEE/CVF Winter Conference on Applications of Computer Vision (WACV), 2025: 1186–1196
\bibitem{48} Bao M, Lyu S, Xu Z, et al. Vision mamba in remote sensing: A comprehensive survey of techniques, applications and outlook. ArXiv preprint arXiv:2505.00630, 2025
\bibitem{49} Fairman F W, Linear control theory: the state space approach. John Wiley \& Sons, 1998
\bibitem{50} Pao H. Forecast of electricity consumption and economic growth in taiwan by state space modeling. Energy, 2009, 34: 1779–1791
\bibitem{51} Johansson R, Robertsson A, Nilsson K, et al. State-space system identification of robot manipulator dynamics, Mechatronics, 2000, 10: 403–418
\bibitem{52} Gu A, Goel K, Gupta A, et al. On the parameterization and initialization of diagonal state space models. Advances in Neural Information Processing Systems, 2022, 35: 35971–35983
\bibitem{53} Gu A, Dao T, Ermon S, et al. Hippo: Recurrent memory with optimal polynomial projections. Advances in neural information processing systems, 2020, 33: 1474–1487
\bibitem{54} Liu Y, Tian Y, Zhao Y, et al. Vmamba: Visual state space model. Advances in neural information processing systems, 2024, 37: 103031–103063
\bibitem{55} Chen K, Chen B, Liu C, et al. Rsmamba: Remote sensing image classification with state space model. IEEE Geoscience and Remote Sensing Letters, 2024
\bibitem{56} Yang J X, Zhou J, Wang J, et al. Hsimamba: Hyperpsectral imaging efficient feature learning with bidirectional state space for classification. ArXiv preprint arXiv:2404.00272, 2024
\bibitem{57} Yao J, Hong D, Li C, et al. Spectralmamba: Efficient mamba for hyperspectral image classification. ArXiv preprint arXiv:2404.08489, 2024
\bibitem{58} Wang G, Zhang X, Peng Z, et al. S ˆ2 mamba: A spatial-spectral state space model for hyperspectral image classification. IEEE Transactions on Geoscience and Remote Sensing, 2025
\bibitem{59} Huang L, Chen Y, He X. Spectral-spatial mamba for hyperspectral image classification. ArXiv preprint arXiv:2404.18401, 2024
\bibitem{60} Zhou W, Kamata S, Wang H, et al. Mamba-in-mamba: Centralized mamba-cross-scan in tokenized mamba model for hyperspectral image classification. Neurocomputing, 2025, 613: 128751
\bibitem{61} He Y, Tu B, Liu B, et al. 3dss-mamba: 3d-spectral-spatial mamba for hyperspectral image classification. IEEE Transactions on Geoscience and Remote Sensing, 2024
\bibitem{62} Shi X, Zhang Y, Liu K, et al. State space models meet transformers for hyperspectral image classification. Signal Processing, 2025, 226: 109669
\bibitem{63} Sheng J, Zhou J, Wang J, et al. Dualmamba: A lightweight spectral-spatial mamba-convolution network for hyperspectral image classification. IEEE Transactions on Geoscience and Remote Sensing, 2024
\bibitem{64} Liao D, Wang Q, Lai T, et al. Joint classification of hyperspectral and lidar data base on mamba. IEEE Transactions on Geoscience and Remote Sensing, 2024
\bibitem{65} Yang A, Li M, Ding Y, et al. Graphmamba: An efficient graph structure learning vision mamba for hyperspectral image classification. IEEE Transactions on Geoscience and Remote Sensing. 2024
\bibitem{66} Ahmad M, Butt M H F, Usama M, et al. Multi-head spatial-spectral mamba for hyperspectral image classification. ArXiv preprint arXiv:2408.01224, 2024
\bibitem{67} Ahmad M, Usama M, Mazzara M, et al. Wavemamba: Spatial-spectral wavelet mamba for hyperspectral image classification. IEEE Geoscience and Remote Sensing Letters, 2024
\bibitem{68} Ahmad M, Butt M H F, Khan A M, et al. Spatial–spectral morphological mamba for hyperspectral image classification. Neurocomputing, 2025, 636: 129995
\bibitem{69} Gao F, Jin X, Zhou X, et al. Msfmamba: Multi-scale feature fusion state space model for multi-source remote sensing image classification. IEEE Transactions on Geoscience and Remote Sensing, 2025
\bibitem{70} Wang C, Huang J, Lv M, et al. A local enhanced mamba network for hyperspectral image classification. International Journal of Applied Earth Observation and Geoinformation, 2024, 133: 104092
\bibitem{71} Chen T, Ye Z, Tan Z, et al. Mim-istd: Mamba-in-mamba for efficient infrared small target detection. IEEE Transactions on Geoscience and Remote Sensing, 2024
\bibitem{72} Ren K, Wu X, Xu L, et al. Remotedet-mamba: A hybrid mamba-cnn network for multi-modal object detection in remote sensing images. ArXiv preprint arXiv:2410.13532, 2024
\bibitem{73} Liu P, Lei S, Li H C. Mamba-moc: A multicategory remote object counting via state space model. ArXiv preprint arXiv:2501.06697, 2025
\bibitem{74} Shen D, Zhu X, Tian J, et al. Htd-mamba: Efficient hyperspectral target detection with pyramid state space model. IEEE Transactions on Geoscience and Remote Sensing, 2025
\bibitem{75} Wang S, Wang C, Shi C, et al. Mask-guided mamba fusion for drone-based visible-infrared vehicle detection. IEEE Transactions on Geoscience and Remote Sensing, 2024
\bibitem{76} Ma X, Zhang X, Pun M O. Rs 3 mamba: Visual state space model for remote sensing image semantic segmentation. IEEE Geoscience and Remote Sensing Letters, 2024
\bibitem{77} Liu M, Dan J, Lu Z, et al. Cm-unet: Hybrid cnn-mamba unet for remote sensing image semantic segmentation. ArXiv preprint arXiv:2405.10530, 2024
\bibitem{78} Wang L, Li D, Dong S, et al. Pyramidmamba: Rethinking pyramid feature fusion with selective space state model for semantic segmentation of remote sensing imagery. ArXiv preprint arXiv:2406.10828, 2024
\bibitem{79} Sun J, Dai Y, Vong C M, et al. Oe-bevseg: An object informed and environment aware multimodal framework for bird’s-eye-view vehicle semantic segmentation. IEEE Transactions on Intelligent Transportation Systems, 2025
\bibitem{80} Ding H, Xia B, Liu W, et al. A novel mamba architecture with a semantic transformer for efficient real-time remote sensing semantic segmentation. Remote Sensing, 2024, 16: 2620
\bibitem{81} Zhu Q, Cai Y, Fang Y, et al. Samba: Semantic segmentation of remotely sensed images with state space model. Heliyon, 2024
\bibitem{82} Zhu E, Chen Z, Wang D, et al. Unetmamba: An efficient unet-like mamba for semantic segmentation of high-resolution remote sensing images. IEEE Geoscience and Remote Sensing Letters, 2024
\bibitem{83} Li L, Chen B, Zou X, et al. Uv-mamba: A dcn-enhanced state space model for urban village boundary identification in high-resolution remote sensing images. ICASSP 2025-2025 IEEE International Conference on Acoustics, Speech and Signal Processing (ICASSP), 2025: 1–5
\bibitem{84} Hu Y, Ma X, Sui J, et al. Ppmamba: A pyramid pooling local auxiliary ssm-based model for remote sensing image semantic segmentation. ArXiv preprint arXiv:2409.06309, 2024
\bibitem{85} Liu C, Chen K, Chen B, et al. Rscama: Remote sensing image change captioning with state space model. IEEE Geoscience and Remote Sensing Letters, 2024
\bibitem{86} Xiao Y, Yuan Q, Jiang K, et al. Frequency-assisted mamba for remote sensing image super-resolution. IEEE Transactions on Multimedia, 2024
\bibitem{87} Huang Y, Miyazaki T, Liu X, et al. Irsrmamba: Infrared image super-resolution via mamba-based wavelet transform feature modulation model. ArXiv preprint arXiv:2405.09873, 2024
\bibitem{88} Wang Y, Yuan W, Xie F, et al. Esatsr: Enhancing super-resolution for satellite remote sensing images with state space model and spatial context. Remote Sensing, 2024: 16: 1956
\bibitem{89} Zhi R, Fan X, Shi J. Mambaformersr: A lightweight model for remote-sensing image super-resolution. IEEE Geoscience and Remote Sensing Letters, 2024
\bibitem{90} Liu Y, Xiao J, Guo Y, et al. Hsidmamba: Exploring bidirectional state-space models for hyperspectral denoising. ArXiv preprint arXiv:2404.09697, 2024
\bibitem{91} Fu G, Xiong F, Lu J, et al. Ssumamba: Spatial-spectral selective state space model for hyperspectral image denoising. IEEE Transactions on Geoscience and Remote Sensing, 2024
\bibitem{92} Zhou H, Wu X, Chen H, et al. Rsdehamba: Lightweight vision mamba for remote sensing satellite image dehazing. ArXiv preprint arXiv:2405.10030, 2024
\bibitem{93} Fu H, Sun G, Li Y, et al. Hdmba: Hyperspectral remote sensing imagery dehazing with state space model. ArXiv preprint arXiv:2406.05700, 2024
\bibitem{94} Liu Z, Chen H, Bai L, et al. Mambads: Near-surface meteorological field downscaling with topography constrained selective state space modeling. IEEE Transactions on Geoscience and Remote Sensing, 2024
\bibitem{95} Debes C, Merentitis A, Heremans R, et al. Hyperspectral and lidar data fusion: Outcome of the 2013 grss data fusion contest. IEEE Journal of Selected Topics in Applied Earth Observations and Remote Sensing, 2014, 7: 2405–2418
\bibitem{96} Rasti B, Hong D, Hang R, et al. Feature extraction for hyperspectral imagery: The evolution from shallow to deep: Overview and toolbox. IEEE Geoscience and Remote Sensing Magazine, 2020, 8: 60–88
\bibitem{97} Chen Y, Jiang H, Li C, et al. Deep feature extraction and classification of hyperspectral images based on convolutional neural networks. IEEE transactions on geoscience and remote sensing, 2016, 54: 6232–6251
\bibitem{98} Hang R, Liu Q, Hong D, et al. Cascaded recurrent neural networks for hyperspectral image classification. IEEE Transactions on Geoscience and Remote Sensing, 2019, 57: 5384–5394
\bibitem{99} Hong D, Gao L, Yao J, et al. Graph convolutional networks for hyperspectral image classification. IEEE Transactions on Geoscience and Remote Sensing, 2020, 59: 5966–5978
\bibitem{100} Vaswani A. Attention is all you need. Advances in Neural Information Processing Systems, 2017
\bibitem{101} Hong D, Han Z, Yao J, et al. Spectralformer: Rethinking hyperspectral image classification with transformers. IEEE Transactions on Geoscience and Remote Sensing, 2021, 60: 1–15
\bibitem{102} Chen K, Chen B, Liu C, et al. Rsmamba: Remote sensing image classification with state space model. IEEE Geoscience and Remote Sensing Letters, 2024
\bibitem{103} He Y, Tu B, Liu B, et al. 3dss-mamba: 3d-spectral-spatial mamba for hyperspectral image classification. IEEE Transactions on Geoscience and Remote Sensing, 2024
\bibitem{104} Yang A, Li M, Ding Y, et al. Graphmamba: An efficient graph structure learning vision mamba for hyperspectral image classification. IEEE Transactions on Geoscience and Remote Sensing, 2024
\bibitem{105} Wang C, Huang J, Lv M, et al. A local enhanced mamba network for hyperspectral image classification. International Journal of Applied Earth Observation and Geoinformation, 2024, 133: 104092
\bibitem{106} Ahmad M, Butt M H F, Khan A M, et al. Spatial–spectral morphological mamba for hyperspectral image classification. Neurocomputing, 2025, 636: 129995
\bibitem{107} Gunther S, Ruthotto L, Schroder J B, et al. Layer-parallel training of deep residual neural networks. SIAM Journal on Mathematics of Data Science, 2020, 2: 1–23
\bibitem{108} Li L, Chen B, Zou X, et al. Uv-mamba: A dcn-enhanced state space model for urban village boundary identification in high-resolution remote sensing images. ICASSP 2025-2025 IEEE International Conference on Acoustics, Speech and Signal Processing (ICASSP), 2025: 1–5
\bibitem{109} Wang J, Zheng Z, Ma A, et al. Loveda: A remote sensing land-cover dataset for domain adaptive semantic segmentation. ArXiv preprint arXiv:2110.08733, 2021
\bibitem{110} Xu J, Xiong Z, Bhattacharyya S P. Pidnet: A real-time semantic segmentation network inspired by pid controllers. Proceedings of the IEEE/CVF conference on computer vision and pattern recognition, 2023: 19529–19539
\bibitem{111} Li R, Zheng S, Zhang C, et al. Multiattention network for semantic segmentation of fine-resolution remote sensing images. IEEE Transactions on Geoscience and Remote Sensing, 2021, 60: 1–13
\bibitem{112} Chen J, Lu Y, Yu Q, et al. Transunet: Transformers make strong encoders for medical image segmentation. ArXiv preprint arXiv:2102.04306, 2021
\bibitem{113} Xu Z, Wu D, Yu C, et al. Sctnet: Single-branch cnn with transformer semantic information for real-time segmentation. Proceedings of the AAAI Conference on Artificial Intelligence, 2024, 38: 6378–6386
\bibitem{114} Weng W, Zhu X. INet: convolutional networks for biomedical image segmentation. IEEE Access. 2021, 9:16591-603
\bibitem{115} Chen T, Ye Z, Tan Z, et al. Mim-istd: Mamba-in-mamba for efficient infrared small target detection. IEEE Transactions on Geoscience and Remote Sensing, 2024
\bibitem{116} Ji S, Wei S, Lu M. Fully convolutional networks for multisource building extraction from an open aerial and satellite imagery data set. IEEE Transactions on geoscience and remote sensing, 2018, 57: 574–586
\bibitem{117} Chen H, Shi Z. A spatial-temporal attention-based method and a new dataset for remote sensing image change detection. Remote Sensing, 2020, 12: 1662
\bibitem{118} Daudt R C, Saux B L and Boulch A. Fully convolutional siamese networks for change detection. 2018 25th IEEE international conference on image processing (ICIP), 2018: 4063–4067
\bibitem{119} Zhang C, Yue P, Tapete D, et al. A deeply supervised image fusion network for change detection in high resolution bi-temporal remote sensing images. ISPRS Journal of Photogrammetry and Remote Sensing, 2020, 166: 183–200
\bibitem{120} Zhang C, Wang L, Cheng S, et al. Swinsunet: Pure transformer network for remote sensing image change detection. IEEE Transactions on Geoscience and Remote Sensing, 2022, 60: 1–13
\bibitem{121} Bandara W G C, Patel V M. A transformer-based siamese network for change detection. IGARSS 2022-2022 IEEE International Geoscience and Remote Sensing Symposium, 2022: 207–210
\bibitem{122} Paranjape J N, Melo C D, Patel V M. A mamba-based siamese network for remote sensing change detection. 2025 IEEE/CVF Winter Conference on Applications of Computer Vision (WACV), 2025: 1186–1196
\bibitem{123} Peng S, Zhu X, Deng H, et al. Fusionmamba: Efficient image fusion with state space model. ArXiv preprint arXiv:2404.07932, 2024
\bibitem{124} Xiao Y, Yuan Q, Jiang K, et al. Frequency-assisted mamba for remote sensing image super-resolution. IEEE Transactions on Multimedia, 2024
\bibitem{125} Huang T, Pei X, You S, et al. Localmamba: Visual state space model with windowed selective scan. European Conference on Computer Vision, 2025: 12–22
\bibitem{126} Chen H, Luo H, Wang C. AfaMamba: Adaptive Feature Aggregation With Visual State Space Model for Remote Sensing Images Semantic Segmentation. IEEE Journal of Selected Topics in Applied Earth Observations and Remote Sensing, 2025
\bibitem{127} Shao M, Tan X, Shang K, et al. A Hybrid Model of State Space Model and Attention for Hyperspectral Image Denoising. IEEE Journal of Selected Topics in Applied Earth Observations and Remote Sensing, 2025
\bibitem{128} Ren Y, Li X, Guo M, et al. Mambacsr: Dual-interleaved scanning for compressed image super-resolution with ssms. ArXiv preprint arXiv:2408.11758, 2024
\bibitem{129} He H, Bai Y, Zhang J, et al. Mambaad: Exploring state space models for multi-class unsupervised anomaly detection. ArXiv preprint arXiv:2404.06564, 2024
\bibitem{130} Jiang X, Han C, Mesgarani N. Dual-path mamba: Short and long-term bidirectional selective structured state space models for speech separation. ICASSP 2025-2025 IEEE International Conference on Acoustics, Speech and Signal Processing (ICASSP), 2025: 1–5
\bibitem{131} Li K, Chen G, Yang R, et al. Spmamba: State-space model is all you need in speech separation. ArXiv preprint arXiv:2404.02063, 2024
\bibitem{132} Verma T, Singh J, Bhartari Y, et al. Soar: Advancements in small body object detection for aerial imagery using state space models and programmable gradients. ArXiv preprint arXiv:2405.01699, 2024
\bibitem{133} Du R, Tang X, Ma J, et al. Mlmamba: A mamba-based efficient network for multi-label remote sensing scene classification. IEEE Transactions on Circuits and Systems for Video Technology, 2025
\bibitem{134} Pan Z, Li C, Plaza A, et al. Hyperspectral image classification with mamba. IEEE Transactions on Geoscience and Remote Sensing, 2024
\bibitem{135} Liu Q, Yue J, Fang Y, et al. Hypermamba: A spectral-spatial adaptive mamba for hyperspectral image classification. IEEE Transactions on Geoscience and Remote Sensing, 2024
\bibitem{136} Huang X, Zhang Y, Luo F, et al. Dynamic token augmentation mamba for cross-scene classification of hyper-spectral image. IEEE Transactions on Geoscience and Remote Sensing, 2024
\bibitem{137} Yan L, Zhang X, Wang K, et al. Contour-enhanced visual state-space model for remote sensing image classification. IEEE Transactions on Geoscience and Remote Sensing, 2024
\bibitem{138} Wang X, Huang Z, Zhang S, et al. Gmsr: Gradient-guided mamba for spectral reconstruction from rgb images. ArXiv preprint arXiv:2405.07777, 2024
\bibitem{139} Wang Y, Liang F, Wang S, et al. Towards an efficient remote sensing image compression network with visual state space model. Remote Sensing, 2025, 17: 425
\bibitem{140} Fu X, Zhang T, Cheng J, et al. Mmr-had: Multi-scale mamba reconstruction network for hyperspectral anomaly detection. IEEE Transactions on Geoscience and Remote Sensing, 2025
\bibitem{141} Shen Y, Xiao L, Chen J, et al. Learning cross-task features with mamba for remote sensing image multi-task prediction. IEEE Transactions on Geoscience and Remote Sensing, 2025
\bibitem{142} Huang J, Yuan X, Lam C T, et al. Lccdmamba: Visual state space model for land cover change detection of vhr remote sensing images. IEEE Journal of Selected Topics in Applied Earth Observations and Remote Sensing, 2025
\bibitem{143} Cao Y, Liu C, Wu Z, et al. Remote sensing image segmentation using vision mamba and multi-scale multi-frequency feature fusion. Remote Sensing, 2025, 17: 1390
\bibitem{144} Zheng J, Fu Y, Chen X, et al. Egcm-unet: Edge guided hybrid cnn-mamba unet for farmland remote sensing image semantic segmentation. Geocarto International, 2025, 40: 2440407
\bibitem{145} Cui X, Zhang L. Mamba-DCAU: state space dual attention center-sampling U-Net for hyperspectral image classification. International Journal of Remote Sensing, 2025, 46: 5523-47
\bibitem{146} Du S, Xing J, Wang S, et al. Stmnet: Scene classification-assisted and texture feature-enhanced multi scale network for large-scale urban informal settlement extraction from remote sensing images. IEEE Journal of Selected Topics in Applied Earth Observations and Remote Sensing, 2024
\bibitem{147} Dong J, Yin H, Li H, et al. Dual hyperspectral mamba for efficient spectral compressive imaging. ArXiv preprint arXiv:2406.00449, 2024
\bibitem{148} Liu Y, Wang X, Shi L, et al. Ls-unet for semantic segmentation of fine-resolution remotely sensed images. 2024 China Automation Congress (CAC), 2024: 5373–5378
\bibitem{149} Li D, Zhao J, Chang C, et al. Lgmamba: Large-scale als point cloud semantic segmentation with local and global state space model. IEEE Geoscience and Remote Sensing Letters, 2024
\bibitem{150} Feng Y, Zhuo L, Zhang H, et al. Hybrid-MambaCD: Hybrid Mamba-CNN Network for Remote Sensing Image Change Detection With Region-Channel Attention Mechanism and Iterative Global-Local Feature Fusion. IEEE Transactions on Geoscience and Remote Sensing, 2025
\bibitem{151} Ye F, Tan S, Huang W, et al. MambaTriNet: A Mamba based Tri-backbone multimodal remote sensing image semantic segmentation model. IEEE Geoscience and Remote Sensing Letters, 2025
\bibitem{152} Li L, Yi J, Fan H, et al. A Lightweight Semantic Segmentation Network Based on Self-attention Mechanism and State Space Model for Efficient Urban Scene Segmentation. IEEE Transactions on Geoscience and Remote Sensing, 2025
\bibitem{153} Wang X, Sun Q, Qi H, et al. Multiscale CNN-Mmaba hybrid network for building change detection in remote sensing images. InThird International Conference on Remote Sensing, Mapping, and Geographic Information Systems (RSMG 2025), 2025, 13791: 247-252
\bibitem{154} Gao R, Xiao Z, Xiong Z. Mamba-based light field super-resolution with efficient subspace scanning. Proceedings of the Asian Conference on Computer Vision, 2024: 531–547
\bibitem{155} Liu C, Cao Y, Jiang D, et al. Efficient vision mamba with dynamic feature fusion for remote sensing image semantic segmentation. 2024 IEEE International Conference on Signal, Information and Data Processing (ICSIDP), 2024: 1–6
\bibitem{156} Hu V T, Baumann S A, Gui M, et al. Zigma: A dit-style zigzag mamba diffusion model. European Conference on Computer Vision, 2024: 149-166
\bibitem{157} Wang C, Huang J, Lv M, et al. A local enhanced mamba network for hyperspectral image classification. International Journal of Applied Earth Observation and Geoinformation, 2024, 133: 104092
\bibitem{158} Cao K, He X, Hu T, et al. Shuffle mamba: State space models with random shuffle for multi-modal image fusion. ArXiv preprint arXiv:2409.01728, 2024
\bibitem{159} Sun J, Dai Y, Vong C M, et al. Oe-bevseg: An object informed and environment aware multimodal framework for bird’s-eye-view vehicle semantic segmentation. IEEE Transactions on Intelligent Transportation Systems, 2025
\bibitem{160} Qiao J, Liao J, Li W, et al. Hi-mamba: Hierarchical mamba for efficient image super-resolution. ArXiv preprint arXiv:2410.10140, 2024
\bibitem{161} Fu G, Xiong F, Lu J, et al. Ssumamba: Spatial-spectral selective state space model for hyperspectral image denoising. IEEE Transactions on Geoscience and Remote Sensing, 2024
\bibitem{162} Wang Q, Fan X, Huang J, et al. Spectral–spatial feature extraction network with ssm–cnn for hyperspectral–multispectral image collaborative classification. IEEE Journal of Selected Topics in Applied Earth Observations and Remote Sensing, 2024
\bibitem{163} Ma J, Li B, Li H, et al. Remote Sensing Change Detection by Pyramid Sequential Processing With Mamba. IEEE Journal of Selected Topics in Applied Earth Observations and Remote Sensing, 2025
\bibitem{164}Sun H, Liu J, Yang J, et al. HMAFNet: Hybrid Mamba-Attention Fusion Network for Remote Sensing Image Semantic Segmentation. IEEE Geoscience and Remote Sensing Letters, 2025
\bibitem{165} Sun H, Yang X, Lu R, et al. Mscnet: Mamba-Based Self-Correction Remote Sensing Change Detection Network. Available at SSRN 5184974, 2025
\bibitem{166} He Y, Tu B, Liu B, et al. HSI-MFormer: Integrating Mamba and Transformer Experts for Hyperspectral Image Classification. IEEE Transactions on Geoscience and Remote Sensing, 2025
\bibitem{167} Wang C, Tsepa O, Ma J, et al. Graph-mamba: Towards long-range graph sequence modeling with selective state spaces. ArXiv preprint arXiv:2402.00789, 2024
\bibitem{168} Li L, Wang H, Zhang W, et al. Stg-mamba: Spatial-temporal graph learning via selective state space model. ArXiv preprint arXiv:2403.12418, 2024
\bibitem{169} Liu C, Lin J, Wang J, et al. Mamba4rec: Towards efficient sequential recommendation with selective state space models. ArXiv preprint arXiv:2403.03900, 2024
\bibitem{170} Yang J, Li Y, Zhao J, et al. Uncovering selective state space model’s capabilities in lifelong sequential recommendation. ArXiv preprint arXiv:2403.16371, 2024
\bibitem{171} Luan X, Fan H, Wang Q, et al. Fmambair: A hybrid state space model and frequency domain for image restoration. IEEE Transactions on Geoscience and Remote Sensing, 2025
\bibitem{172} Zhuang P, Zhang X, Wang H, et al. Fahm: Frequency-aware hierarchical mamba for hyperspectral image classification. IEEE Journal of Selected Topics in Applied Earth Observations and Remote Sensing, 2025
\bibitem{173} Wu L, Peng J, Yang B, et al. Ssmif: Enhanced spatial-spectral mamba interactive fusion network for hyperspectral change detection. 2024 IEEE International Conference on Signal, Information and Data Processing (ICSIDP), 2024: 1–5
\bibitem{174} Ming R, Chen N, Peng J, et al. Semantic tokenization-based mamba for hyperspectral image classification. IEEE Journal of Selected Topics in Applied Earth Observations and Remote Sensing, 2025
\bibitem{175} Zhang G, Zhang Z, Deng J, et al. S 2 crossmamba: Spatial–spectral cross-mamba for multimodal remote sensing image classification. IEEE Geoscience and Remote Sensing Letters, 2024
\bibitem{176} Li J, Liu Z, Liu S, et al. Mbssnet: A mamba-based joint semantic segmentation network for optical and sar images. IEEE Geoscience and Remote Sensing Letters, 2025
\bibitem{177} Zhu C, Deng S, Song X, et al. Mamba collaborative implicit neural representation for hyperspectral and multispectral remote sensing image fusion. IEEE Transactions on Geoscience and Remote Sensing, 2025
\bibitem{178} He Y, Tu B, Jiang P, et al. Igroupss-mamba: Interval group spatial-spectral mamba for hyperspectral image classification. IEEE Transactions on Geoscience and Remote Sensing, 2024
\bibitem{179} Yue Z, Xu J, Yan Y, et al. TFFNet: Transform Fusion Fuzzy Network for Multimodal Remote Sensing Classification. IEEE Geoscience and Remote Sensing Letters, 2025
\bibitem{180} Wang R, Zhang J, Lu X, et al. JM-Guided Sentinel 1/2 Fusion and Lightweight APM-UNet for High-Resolution Soybean Mapping. Remote Sensing, 2025, 17: 3934
\bibitem{181} Li Z, Zhao L, Lu Y, et al. Mamba for Remote Sensing: Architectures, Hybrid Paradigms, and Future Directions, 2025
\bibitem{182} Chen B, Chen K, Yang Y, et al. SeG-SR: Integrating Semantic Knowledge Into Remote Sensing Image Super-Resolution via Vision-Language Model. IEEE Transactions on Geoscience and Remote Sensing, 2025, 63: 1-15
\bibitem{183} Ma H, Zhang C, Bian Y, et al. Fairness-guided few-shot prompting for large language models. Advances in Neural Information Processing Systems, 2023, 36: 43136–43155
\bibitem{184} Jiang M, Bao K, Zhang J, et al. Item-side fairness of large language model-based recommendation system. Proceedings of the ACM on Web Conference 2024, 2024: 4717–4726
\bibitem{185} Li D, Bhatti U A, Huang M, et al. Hyperfusionmamba: Hybrid spectral-spatial fusion mamba for efficient hyperspectral image classification with few-shot learning. Available at SSRN 5224711, 2025
\bibitem{186} Cheng G, Huang Y, Li X, et al. Change detection methods for remote sensing in the last decade: A comprehensive review. Remote Sensing, 2024, 16: 2355
\bibitem{187} Chen W, Gao F, Gan Y, et al. Wavelet-Assisted Mamba for Satellite-Derived Sea Surface Temperature Super-Resolution. IEEE Transactions on Geoscience and Remote Sensing, 2025
\bibitem{188} Chen K, Liu C, Chen B, et al. Dynamicvis: An efficient and general visual foundation model for remote sensing image understanding. ArXiv preprint arXiv:2503.16426, 2025
\bibitem{189} Ma X, Wu Q, Zhao X, et al. Sam-assisted remote sensing imagery semantic segmentation with object and boundary constraints. IEEE Transactions on Geoscience and Remote Sensing, 2024
\bibitem{190} Kirillov A, Mintun E, Ravi N, et al. Segment anything. Proceedings of the IEEE/CVF International Conference on Computer Vision, 2023: 4015–4026
\bibitem{191} Girdhar R, El-Nouby A, Liu Z, et al. Imagebind: One embedding space to bind them all. Proceedings of the IEEE/CVF Conference on Computer Vision and Pattern Recognition, 2023: 15180–15190
\bibitem{192} Wang X, Wang W, Cao Y, et al. Images speak in images: A generalist painter for in-context visual learning. Proceedings of the IEEE/CVF Conference on Computer Vision and Pattern Recognition, 2023: 6830–6839
\bibitem{193} Ma X, Xu X, Zhang X, et al. Adjacent-scale multimodal fusion networks for semantic segmentation of remote sensing data. IEEE Journal of Selected Topics in Applied Earth Observations and Remote Sensing, 2024
\bibitem{194} L¨uddecke T, Ecker A. Image segmentation using text and image prompts. Proceedings of the IEEE/CVF conference on computer vision and pattern recognition, 2022: 7086–7096
\bibitem{195} Chen X, Wang X, Changpinyo S, et al. Pali: A jointly-scaled multilingual language-image model. ArXiv preprint arXiv:2209.06794, 2022
\bibitem{196} Yang J, Jin H, Tang R, et al. Harnessing the power of llms in practice: A survey on chatgpt and beyond. ACM Transactions on Knowledge Discovery from Data, 2024, 18: 1–32
\bibitem{197} Radford A, Kim J W, Hallacy C, et al. Learning transferable visual models from natural language supervision. International conference on machine learning, 2021: 8748–8763
\bibitem{198} Yang J, Liu T, Zhang L, et al. Effective road segmentation with selective state space model and frequency feature compensation. IEEE Transactions on Geoscience and Remote Sensing, 2024
\bibitem{199} Li Y, Ren J, Fu H, et al. Gassm: Global attention and state space model based end-to-end hyperspectral change detection. Journal of the Franklin Institute, 2025, 362: 107424
\bibitem{200} Ma X, Zhang X, Ding X, et al. Decomposition-based unsupervised domain adaptation for remote sensing image semantic segmentation. IEEE Transactions on Geoscience and Remote Sensing, 2024
\bibitem{201} Chen K, Liu C, Chen B, et al. RSRefSeg 2: Decoupling Referring Remote Sensing Image Segmentation with Foundation Models. ArXiv preprint arXiv:2507.06231, 2025
\bibitem{202} Chen B, Liu L, Zou Z, et al. Target detection in hyperspectral remote sensing image: Current status and challenges. Remote Sensing, 2023, 15: 3223
\bibitem{203} Chen B, Chen K, Yang M, et al. Heterogeneous mixture of experts for remote sensing image super-resolution. IEEE Geoscience and Remote Sensing Letters, 2025
\bibitem{204} Otter D W, Medina J R, Kalita J K. A survey of the usages of deep learning for natural language processing. IEEE transactions on neural networks and learning systems, 2020, 32: 604–624
\bibitem{205} Shi X, Zhang Y, Liu K, et al. State space models meet transformers for hyperspectral image classification, Signal Processing, 2025, 226: 109669
\bibitem{206} Duc C M, Fukui H. Satmamba: Development of foundation models for remote sensing imagery using state space models. ArXiv preprint arXiv:2502.00435, 2025
\bibitem{207} Devalal S, Karthikeyan A. Lora technology-an overview, in 2018 second international conference on electronics. Communication and aerospace technology (ICECA), 2018: 284–290
\bibitem{208} Sundaram J P S, Du W, Zhao Z. A survey on lora networking: Research problems, current solutions, and open issues. IEEE Communications Surveys \& Tutorials, 2019, 22: 371–388
\bibitem{209} Chen Z, Duan Y, Wang W, et al. Vision transformer adapter for dense predictions. ArXiv preprint arXiv:2205.08534, 2022
\bibitem{210} Xing Z, Ye T, Yang Y, et al. Segmamba: Long-range sequential modeling mamba for 3d medical image segmentation. International Conference on Medical Image Computing and Computer-Assisted Intervention, 2024: 578–588
\bibitem{211} Wang C, Zheng W, Huang Y, et al. V2m: Visual 2-dimensional mamba for image representation learning. ArXiv preprint arXiv:2410.10382, 2024
\bibitem{212} Wei R, Xu S, Zhong L, et al. Lightmamba: Efficient mamba acceleration on fpga with quantization and hardware co-design, 2025 Design, Automation \& Test in Europe Conference (DATE), 2025: 1–7
\bibitem{213} Mattern J, Hohr K. Mamba-chat, 2023
\bibitem{214} Hu E J, Shen Y, Wallis P, et al. Lora: Low-rank adaptation of large language models. ArXiv preprint arXiv:2106.09685, 2021
\bibitem{215} Wang X, Wang Z, Jiang J, et al. Treat stillness with movement: Remote sensing change detection via coarse-grained temporal foregrounds mining. ArXiv preprint arXiv:2408.08078, 2024
\bibitem{216} Chen Z, Shamsabadi E A, Jiang S, et al. Vision mamba-based autonomous crack segmentation on concrete, asphalt, and masonry surfaces. ArXiv preprint arXiv:2406.16518, 2024
\bibitem{217} Fu D Y, Epstein E L, Nguyen E, et al. Simple hardware-efficient long convolutions for sequence modeling. International Conference on Machine Learning, 2023: 10373–10391
\bibitem{218} Touvron H, Lavril T, Izacard G, et al. Llama: Open and efficient foundation language models. ArXiv preprint arXiv:2302.13971, 2023
\bibitem{219} Brown T B. Language models are few-shot learners. ArXiv preprint arXiv:2005.14165, 2020
\bibitem{220} Houlsby N, Giurgiu A, Jastrzebski S, et al. Parameter efficient transfer learning for nlp. International conference on machine learning, 2019: 2790–2799
\bibitem{221} Lin Z, Madotto A, Fung P. Exploring versatile generative language model via parameter-efficient transfer learning. ArXiv preprint arXiv:2004.03829, 2020
\bibitem{222} Kojima T, Gu S S, Reid M, et al. Large language models are zero-shot reasoners. Advances in neural information processing systems, 2022, 35: 22199–22213
\bibitem{223} Xiao G, Lin J, Seznec M, et al. Smoothquant: Accurate and efficient post-training quantization for large language models. International Conference on Machine Learning, 2023: 38087–38099
\bibitem{224} Zhang H, Guo H, Chen K, et al. FoBa: A Foreground–Background Co-Guided Method and New Benchmark for Remote Sensing Semantic Change Detection, IEEE Transactions on Geoscience and Remote Sensing, 2025, 63: 1-19
\bibitem{225} Ma C, Ji Y, Ye J, et al. Towards interpretable counterfactual generation via multimodal autoregression. ArXiv preprint arXiv:2503.23149, 2025
\bibitem{226} Rao V M, Hla M, Moor M, et al. Multimodal generative ai for medical image interpretation. Nature, 2025, 639: 888–896
\bibitem{227} Hu Y, Wang L, Liu X, et al. Simulating the real world: A unified survey of multimodal generative models. ArXiv preprint arXiv:2503.04641, 2025
\bibitem{228} Li J, Zhang C, Zhu W, et al. A comprehensive survey of image generation models based on deep learning. Annals of Data Science, 2025, 12: 141–170
\bibitem{229} Wei H, Sun Y, Li Y. DeepSeek-OCR: Contexts Optical Compression. ArXiv preprint arXiv:2510.18234, 2025
\bibitem{230} Duan Y, Che L, Cheng X, et al. EMambaEdgeNet: Featuring Feature Decoupling and Geometry-Aware Fusion with Mamba Architecture. InInternational Conference on Neural Information Processing, 2025: 517-531







\end{thebibliography}
\end{document}